\documentclass{article}

\usepackage[utf8]{inputenc} 
\usepackage[T1]{fontenc}    
\usepackage{graphicx}
\usepackage{tabularx}

\usepackage{amsmath}
\usepackage{mathtools}

\usepackage{float}
\usepackage{hyphenat} 
\usepackage{arydshln} 
\usepackage{booktabs} 
\usepackage{multicol}
\usepackage{stackengine}
\usepackage{booktabs,multirow}
\usepackage{url}
\usepackage{hyperref}
\usepackage{mathtools}
\usepackage{subfigure}
\usepackage{verbatim}

\usepackage{xcolor}
\definecolor{darkred}{rgb}{0.8,0,0}
\definecolor{darkgreen}{rgb}{0,0.5,0}
\definecolor{darkblue}{rgb}{0,0,0.7}
\definecolor{darkpurple}{rgb}{0.4,0,0.6}
\definecolor{lightgray}{rgb}{0.92,0.92,0.92}
\definecolor{lightpink}{rgb}{1.00,0.90,0.90}
\newcommand{\red}[1]{\noindent{\color{red}{#1}}}
\newcommand{\blue}[1]{\noindent{\color{blue}{#1}}}
\newcommand{\darkgreen}[1]{\noindent{\color{darkgreen}{#1}}}
\definecolor{royalblue}{rgb}{0.25, 0.41, 0.88}

\hypersetup{
  colorlinks,
  citecolor=royalblue,
  linkcolor=royalblue,
  urlcolor=royalblue
}

\usepackage[preprint, nonatbib]{neurips_2024}
\usepackage[numbers]{natbib}

\title{Inferring Underwater Topography with FINN}

\author{
Coşku Can Horuz\textsuperscript{$2$}, Matthias Karlbauer\textsuperscript{$1$}, \\
\textbf{Timothy Praditia\textsuperscript{$3$}, Sergey Oladyshkin\textsuperscript{$3$}, Wolfgang Nowak\textsuperscript{$3$}, Sebastian Otte\textsuperscript{$2$}
}
 \vspace{1.3mm}\\
\normalfont
\textsuperscript{$1$}Neuro-Cognitive Modeling, University of T{\"u}bingen,\\
\textsuperscript{$2$}Adaptive AI Lab, Institute of Robotics and Cognitive Systems, University of L{\"u}beck, \\
\textsuperscript{$3$}Department of Stochastic Simulation and Safety Research for Hydrosystems, University of Stuttgart.
}

\begin{document}

\maketitle

\begin{abstract}
  Spatiotemporal partial differential equations (PDEs) find extensive application across various scientific and engineering fields. While numerous models have emerged from both physics and machine learning (ML) communities, there is a growing trend towards integrating these approaches to develop hybrid architectures known as physics-aware machine learning models. Among these, the finite volume neural network (FINN) has emerged as a recent addition. FINN has proven to be particularly efficient in uncovering latent structures in data. In this study, we explore the capabilities of FINN in tackling the shallow-water equations, which simulates wave dynamics in coastal regions. Specifically, we investigate FINN's efficacy to reconstruct underwater topography based on these particular wave equations. Our findings reveal that FINN exhibits a remarkable capacity to infer topography solely from wave dynamics, distinguishing itself from both conventional ML and physics-aware ML models. Our results underscore the potential of FINN in advancing our understanding of spatiotemporal phenomena and enhancing parametrization capabilities in related domains.
\end{abstract}

\section{Introduction}
\label{sec:introduction}
Machine learning models that incorporate physical principles follow the governing rules about the problem at hand. Such a physics-aware model outperform pure ML models when tackling physical phenomena \cite{karniadakis2021physics,raissi2019physics}. The success and enhanced performance of physics-aware ML models have been demonstrated across a variety of applications, underscoring their power \cite{guen2020disentangling,li2020fourier,long2018pde,seo2019physics,sitzmann2020implicit}. Nevertheless, as in every emerging field, they have not reached their full potential. Especially the physics-aware models lack the ability to structurally incorporate explicitly defined physical equations as well as the capacity to reconstruct underlying physical structures from the data. Recent introduction of the finite volume neural network (FINN) marks a significant step forward in addressing these limitations \cite{praditia2021finite,praditia2022learning,karlbauer2021composing}. FINN combines the well-studied physical models with the learning capabilities of artificial neural networks and models partial differential equations (PDEs) in a mathematically compositional manner. A notable feature of FINN is its ability to directly incorporate boundary conditions (BC) demonstrating an example of explicitly embedded physical structure \cite{karlbauer2021composing}. Excitingly, FINN can also handle the BC that were not considered during training phase and infer unknown Dirichlet boundary conditions, which is an example of hidden information inference from data \cite{Horuz2022,Horuz2023}.

In this study, we explore the FINN concept to reconstruct underwater topography using the shallow-water equations (SWE), a set of PDEs that model wave dynamics. By adapting FINN to solve these equations, we open new pathway for a broader area of application of the model. We compare the quality of the parameterized topography with two models, DISTANA \cite{karlbauer2019distributed} and PhyDNet \cite{guen2020disentangling}. Our results indicate that the models are able to reconstruct the physical processes and predict the wave patterns, but FINN's embedded robust physical structure makes it stand out.

\section{Models}
\label{sec:models}
This section provides an overview of the models under comparison: PhyDNet, DISTANA, and FINN. The descriptions of DISTANA and PhyDNet are presented with less detail in comparison to FINN, as the model of interest.

\subsection{FINN}
The finite volume neural network (FINN) introduced in \cite{karlbauer2021composing,praditia2021finite} represents a novel fusion of the finite volume method (FVM), a well-established numerical technique in computational physics, with the learning capabilities of deep neural networks. The FVM spatially discretizes continuous PDEs, transforming them into algebraic (linear) equations defined over a finite number of control volumes. These control volumes possess distinct states and exchange fluxes following the conservation laws. By embedding physical principles within the FVM structure, FINN is inherently constrained to enforce (partially) known laws of physics. Consequently, this hybrid approach yields an interpretable, highly generalizable, and robust computational methodology.

FINN solves PDEs that convey non-linear spatiotemporal advection-diffusion-reaction processes. It solves the equations in the following form (as formulated in \cite{karlbauer2021composing}):
\begin{equation}
    \label{eq:general_pde}
    \frac{\partial u}{\partial t} = D(u) \frac{\partial^2 u}{\partial x^2} - v(u)\frac{\partial u}{\partial x} + q(u),
\end{equation}
where $u$ is the unknown function of time $t$ and spatial coordinate $x$, which encodes a state. The objective of a PDE solver (if the PDE was fully known) is to find the value of $u$ in all time steps and spatial locations. However, \autoref{eq:general_pde} is composed of three, often unknown functions, which modify $u$, i.e. $D$, $v$, and $q$. $D$ is the diffusion coefficient, which manages the equilibration between high and low concentrations, $v$ is the advection velocity, which represents the movement of concentration due to the bulk motion of a fluid, and $q$ is the source/sink term, which increases or decreases the quantity of $u$ locally. FINN is capable to model the physical systems corresponding to \autoref{eq:general_pde} in both one and two dimensional problems.

\begin{figure}[t!]
    \centering
    \includegraphics[width=\textwidth]{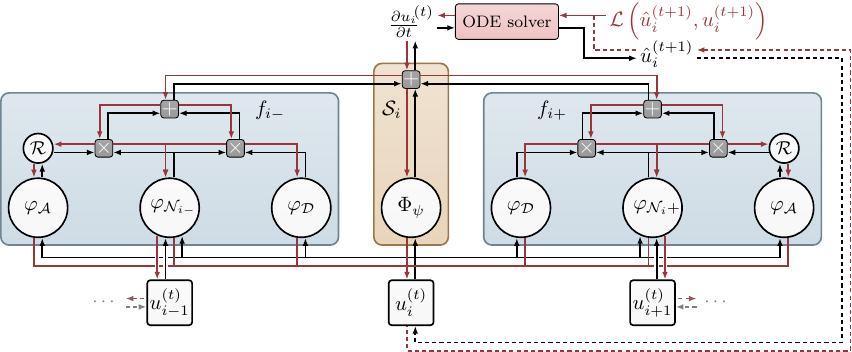}
    \caption{\textit{FINN architecture}. The composition of the modules to solve an advection-diffusion PDE for one control volume in FVM mesh with the help of adjacent volumes. Black lines show the forward pass, whereas red lines indicate the gradient flow. Figure from \cite{karlbauer2021composing}.}
    \label{fig:finn}
\end{figure}

\autoref{fig:finn} and \autoref{eq:general_pde_volume} illustrate how FINN approach models the PDE for a single control volume $i$. The first- and second-order spatial derivatives $\smash{\left(\frac{\partial u}{\partial x}, \smash{\frac{\partial^2 u}{\partial x^2}}\right)}$, are approximated by a linear layer, $\varphi_{\mathcal{N}}$, aiming to learn the FVM stencil, i.e., the exchange terms between adjacent volumes. $\varphi_\mathcal{D}$, $\varphi_\mathcal{A}$, $\Phi_\psi$ are also linear layers approximating $D(u), v(u)$ and $q(u)$, respectively. $\mathcal{R}(\cdot)$ is a point-wise mapping, which manipulates the ReLU activation function to ensure the correct flux direction (i.e. the sign of advective values). The estimated values are combined together as in \autoref{eq:general_pde} in two different kernels, called flux and state kernel. Flux kernel computes the values in the adjacent volumes wheres state kernel receives these fluxes and combines them with the source/sink term which simply decides if there is an extra incoming or outgoing flux in that particular volume.

\begin{figure}[t!]
\begin{equation}\label{eq:general_pde_volume}
\begin{aligned}\frac{\partial u_i}{\partial t} & = \underbracket{\underbracket{\overbracket{\addstackgap[17.2pt]\phantom{}D\left(u_i\right)}^{\displaystyle\varphi_\mathcal{D}} \overbracket{\addstackgap[17.2pt]\phantom{}\frac{\partial^2 u_i}{\partial x^2}}^{
	\displaystyle
	\substack{
	\displaystyle\varphi_{\mathcal{N}_{i-}}\\
	\displaystyle+\\
	\displaystyle\varphi_{\mathcal{N}_{i+}}
	}
	} - \overbracket{\addstackgap[17.2pt]\phantom{}v\left(u_i\right)}^{\displaystyle\mathcal{R}(\varphi_\mathcal{A})} \overbracket{\addstackgap[17.2pt]\phantom{}\frac{\partial u_i}{\partial x}}^{
	\displaystyle
 	\substack{
 	\vspace{1pt}
 	\displaystyle\varphi_{\mathcal{N}_{i-}}\\
 	\vspace{1pt}
 	\texttt{\normalsize or}\\
 	\displaystyle\varphi_{\mathcal{N}_{i+}}
	}
	}
	}_{\displaystyle\mathcal{F}_i = f_{i-} + f_{i+}} + \overbracket{\addstackgap[17.2pt]\phantom{}q\left(u_i\right)}^{\displaystyle\Phi_\psi}}_{\displaystyle\mathcal{S}_i}\end{aligned}
\end{equation}
\vspace{-0.5cm}
\end{figure}

As can be seen in \autoref{fig:finn}, all modules' outputs are summed up to conclude in $\smash{\frac{\partial u}{\partial t}}$, which results in a system of ordinary differential equation (ODE) solvable over time, e.g. via Euler method. On a high level, FINN operates as a data-driven ODE maker, utilizing spatial data (PDE response or measurements) to resolve the spatial dimensions through FVM, thereby reducing the problem to a temporal ODE. The resulting ODE is fed into a solver, which produces a prediction for the next time step. This process is applied to all control volumes in the FVM mesh and the prediction is fed back to the model in a closed-loop fashion until the end of the sequence is reached. The error is computed via $\smash{\mathcal{L}\left(\hat{u}^{(1:T)}_i, u^{(1:T)}_i\right)}$, where $(1:T)$ corresponds to the entire sequence and $i$ is the discretized spatial control volume index. Mean squared error is used as loss function. For a comprehensive understanding of FINN's underlying mechanisms, including its kernel functions and different neural network structures, readers are referred to \cite{karlbauer2021composing}.

\subsection{DISTANA}\label{sec:distana}
The distributed spatiotemporal graph artificial neural network architecture (DISTANA) is a hidden state inference model for time series prediction. It encodes two different kernels in a graph structure. First, the \textit{prediction kernel} (PK) network predicts the dynamics at each spatial position while being applied to each node of the underlying mesh. Second, the \textit{transition kernel} (TK) network coordinates the lateral information flow between PKs. Thus allowing the model to process spatiotemporal data. The PKs consist of feed-forward neural networks combined with LSTM units \cite{hochreiter1997long}. In this work, similar to \cite{karlbauer2019distributed}, linear mappings were used as TKs since the data is represented on a regular grid and does not require more complex processing of lateral information.
All PKs and TKs share weights. DISTANA has been shown to perform better compared to convolutional neural networks, recurrent neural networks, convolutional LSTM, and similar models \cite{karlbauer2019distributed,karlbauer2021latent}. The model's performance and its applicability to spatiotemporal data makes it a suitable pure ML baseline for comparison.

\subsection{PhyDNet}\label{sec:phydnet}
PhyDNet is a physics-aware encoder-decoder model introduced in \cite{guen2020disentangling}. First, it encodes the input at time step $t$. Afterward, the encoded information is disentangled into two separate networks, PhyCell and ConvLSTM-Cell. Inspired by physics, PhyCell implements spatial derivatives up to a desired order and can approximate solutions of a wide range of PDEs, e.g., heat equation, wave equation, and the advection-diffusion equation. Moreover, the model covers the residual information that is not subsumed by the physical norms using neural networks. Concretely, the ConvLSTM-Cell complements the PhyCell and approximates the residual information in a convolutional deep learning fashion. The outputs of the networks are combined and fed into a decoder in order to generate a prediction of the unknown function at the next time step. PhyDNet is a state-of-the-art physics-aware neural network, which applies to advection-diffusion equations and was therefore selected as a physics-aware baseline in this study.
\section{Shallow-Waters Equations}
\label{sec:swe}
\begin{figure}[ht]
    \centering
    \includegraphics[scale=2.5]{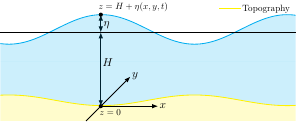}
    \caption{\textit{Illustration of SWE.} $x$, $y$ and $z$ correspond to width, length and depth, respectively. In the case of flat topography, $H$ is equal in every location. However, a non-flat topography is used in this study and $H$ has different values in different locations. $\eta$ is the deviance of the free surface (thick blue line) from the average depth (thick black line) and it is the unknown state. In topography reconstruction, $H$ is reproduced from $\eta$.}
    \label{fig:swe}
\end{figure}
Shallow water equations (SWE) represent fluid dynamics where the wavelength $\lambda$ (distance between consecutive waves) is much larger than the depth of the fluid $H$:
\begin{equation}
    \label{eq:swe_assumption}
    \frac{H}{\lambda} \ll 1
\end{equation}
The corresponding fluid does not have to be water but SWE are predominantly applied to aquatic environments. This is due to the common occurrence of the shallow-water assumption given in \autoref{eq:swe_assumption} in modelling the dynamics of stream beds, such as those found in rivers and seas. SWE are essentially a simplified version of the Navier-Stokes equations, derived from the principles of continuity and momentum conservation, specifically tailored to account for the properties of water. In this study, the nonlinear form of the continuity equation is considered, whereas momentum equations are taken to be linear. The specific form of the SWE employed in this study are taken from \cite{kundu2004} and derived in \autoref{appendix:swe_derivation}. SWE are defined as follows (see \autoref{fig:swe} for an illustration of SWE with the respective terms from the equations below):
\begin{align}
    &\frac{\partial \eta}{\partial t} + \frac{\partial}{\partial x}[u(H + \eta)] + \frac{\partial}{\partial y}[v(H + \eta)] = 0, \label{eq:swe_mom}\\
    &\frac{\partial u}{\partial t} = -g\frac{\partial \eta}{\partial x}, \label{eq:swe_cont1}\\
    &\frac{\partial v}{\partial t} = -g\frac{\partial \eta}{\partial y}\label{eq:swe_cont2}
\end{align}
where $u$ (not to confuse with the unknown state in Equations \ref{eq:general_pde} and \ref{eq:general_pde_volume}) and $v$ are the velocity vectors in $x$- and $y$-directions, respectively. $H$ is the \textit{topography} of the fluid represented by the depth length of each grid in meters. The term topography is interchangeably used with depth highlighting that $H$ represents the depth in case of a flat surface whereas it represents a topography when it is not flat. $\eta$ is the displacement of the free surface, essentially capturing the vertical deviation of the fluid surface from the mean depth as waves propagate. Lastly, $g$ expresses the gravitational force that arises from the gravitational field of the earth. A thorough derivation and the visualization of SWE are provided in \autoref{appendix:swe_derivation}.
\section{Underwater Topography Reconstruction}\label{sec:und_top_rec}
\begin{figure}[t]
    \centering
    \includegraphics[width=0.7\textwidth]{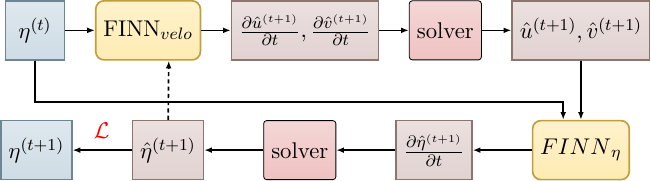}
    \caption{Modification of FINN to solve SWE. Dark black arrows show the forward pass. Loss is computed between $\eta^{(t+1)}$ and $\hat{\eta}^{(t+1)}$ as the mean squared error. The prediction is fed back into the model in a closed-loop fashion (dashed arrow).}
    \label{fig:finn_swe}
\end{figure}

Following the previous work in \cite{Horuz2023}, this study delves into FINN's capacity to infer hidden information in data. Nevertheless, there are substantial differences. First of all, SWE is two-dimensional and is a combination of three equations. Second, the aim is to reconstruct the topography from $\eta$ but $H$ does not have a direct influence on $\eta$. It also does not change the physical domain drastically as in the case of boundary conditions such that a change in $H$ causes only a small and indirect change in $\eta$ through velocities. Therefore, it would be fair to say that the topography $H$ in SWE is literally \textit{the underlying structure} in data. It is closely coupled to the velocities. The relationship between velocity and topography is mathematically expressed through the eigenvalues of the velocity tensors $u \pm \sqrt{g \cdot H}$ and $v \pm \sqrt{g \cdot H}$ (cf. \cite{cfl2003}). Through this tied relation between velocity and topography, it is possible for FINN to infer $H$. As SWE is different than the other equations in FINN's repertoire, a corresponding model modification was necessary. The visualization of this modification can be seen in \autoref{fig:finn_swe}. Each FINN module, corresponds to the architecture in \autoref{fig:finn} without the solver (it is specifically given in \autoref{fig:finn_swe}). Different FINN modules approximate different unknowns (i.e. $\eta$ and the velocities $u$ and $v$).

The adaptation process involves sequentially computing the unknown state $\eta$ in \autoref{eq:swe_mom} and also the velocities $u$ and $v$. $\eta^{(t=0)}$ is the initial condition which is fed into the model as in the standard FINN way. This first model, called \textit{FINN$_{velo}$}, receives the initial state as input. For the initial state, the following Gaussian bump is used:
\begin{equation}\label{eq:gauss}
    \eta^{(t=0)}_{x,y} = \operatorname{exp}\left(-\left( \frac{(x-x_{0})^2}{2\sigma^2} + \frac{(y-y_{0})^2}{2\sigma^2} \right)\right)
\end{equation}
$x_0$ and $y_0$ are randomly chosen points on the grid for each sequence and standard deviation $\sigma$ is set to $5\cdot 10^{4}$. The Gaussian bump simulates the effect of a huge rock dropped into the water to create waves. Depending on the initial state, FINN$_{velo}$ approximates the time derivatives of the velocities at the next time step $(t+1)$ and produces the ODEs $\smash{\frac{\partial \hat{u}^{(t+1)}}{\partial t}}$ and $\smash{\frac{\partial \hat{v}^{(t+1)}}{\partial t}}$. These predictions are sent to the solver to compute $\hat{u}^{(t+1)}$ and $\hat{v}^{(t+1)}$. Euler solver is used in our study, although it is possible to use finer-grained solvers. In the preliminary studies, the Euler method worked sufficiently well and thus the method with the least computational cost is chosen. The resulting velocities for the next step along with $\eta^{(t)}$ are provided to the second model, which is called \textit{FINN$_{\eta}$}. To estimate $\frac{\partial \hat{\eta}^{(t+1)}}{\partial t}$, the velocities $u$ and $v$ evaluated at $(H_{i,j} + \hat{\eta}^{(t)}_{i,j})$, where $i$ and $j$ correspond to the spatial coordinates, must be approximated (cf. \autoref{eq:swe_mom}). These intermediate values are computed as a combination of the neural network predictions inside FINN$_{\eta}$. Afterwards the spatial derivative of the intermediate values is determined via the finite difference method (FDM), a simplified version of FVM. This procedure results in $\smash{\frac{\partial \hat{\eta}^{(t+1)}}{\partial t}}$, which is again an ODE that is solved by the solver. The resulting state $\hat{\eta}^{(t+1)}$ is fed back into FINN$_{velo}$ in a closed-loop fashion until the end of the sequence is reached.

The above mentioned adaptation of FINN to solve multiple equations necessitated significant changes, not only in its operational framework but also in its neural network architecture. In standard FINN, $\varphi_{\mathcal{N}_{i-}}$ and $\varphi_{\mathcal{N}_{i+}}$ approximate the stencils between two adjacent volumes. This allows FINN to be adapted to both unstructured and structured grids. In current work, $\varphi_{\mathcal{N}}$ estimates the spatial derivatives such as $-g\frac{\partial \hat{\eta}}{\partial x}$ and $-g\frac{\partial \hat{\eta}}{\partial y}$ in continuity equations which are equal to the time derivatives $\frac{\partial u}{\partial t}$ and $\frac{\partial v}{\partial t}$, respectively. $g$ is the gravitational constant and the network weights can effectively generalize for $g$ alongside the spatial derivatives. These derivatives are approximated by multilayer perceptrons (MLP) with an extra inductive bias. It is implemented in MLP through the explicit presentation of adjacent cells similar to FVM/FDM. The tensors $\eta_{i,j}^{(t)}$ and $\eta_{i+1,j}^{(t)}$ $\forall i \le (X-1)$, where $X$ corresponds to the width of the domain, are stacked on top of each other and provided to the model jointly. Accordingly, tensors $\eta_{i,j}^{(t)}$ and $\eta_{i,j+1}^{(t)}$ used for the horizontal direction in the same way. Thus the network receives each time two horizontally or vertically adjacent cells as input. This configuration enables the network to adjust its weights according to the value changes between adjacent cells. An important point worth considering in such implementation of feedforward neural networks is the utilization of boundary conditions. Specifically, when the cells at locations $(i, j)$ and $(i{+}1, j)$ are stacked, it is not possible to take the last cell on the edge as $i$ because the cell $(i{+}1)$ would simply not exist. Therefore $i$ has to be less than or equal to $(X - 1)$. This means that when two adjacent cells are stacked together, the resulting tensor will have a size that is one unit smaller than the spatial domain size. Hence, when adjacent cells are stacked together, the boundaries should be handled carefully. In this study, the boundary conditions are applied to the output of the networks explicitly. For velocity tensors, boundary values are set to zero to adhere to the no-slip condition, a fundamental principle in fluid dynamics. For $\eta$-intermediate tensors, which are estimated in FINN$_\eta$ module, the boundary values of $\eta$ are used such that the value leaving the domain from one side is fed back into the domain in the same side again. This is done to create the wall effect on the edges.

Changes have also been made to the comparative models in order to reconstruct the topography. In DISTANA, lateral information is shared across the prediction kernel on the domain. This is accomplished by convolution kernels applied to data (the unknown state). To embed the topography in the model, we stacked the topography grid onto data, allowing kernels to adjust themselves via gradient descent. In the convolutional sense, basically, another extra input channel was added through stacking operation. PhyDNet adopts a similar strategy, with the primary implementation being in the ConvLSTM encoder branch of the model.

\begin{figure}[t!]
    \subfigure[Train]{\includegraphics[width=0.5\textwidth]{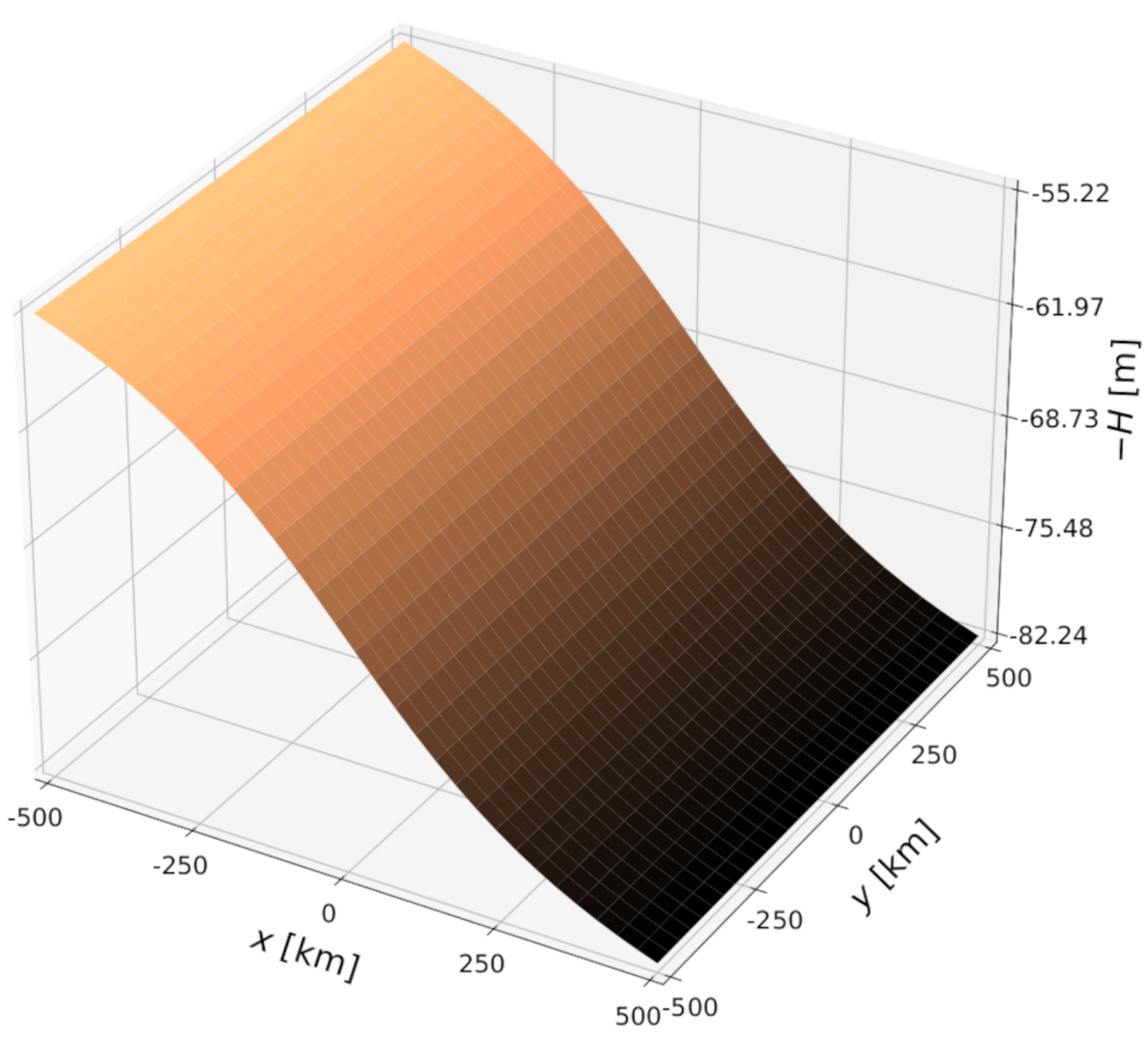}}
    \hfill
    \subfigure[Inference]{\includegraphics[width=0.48\textwidth]{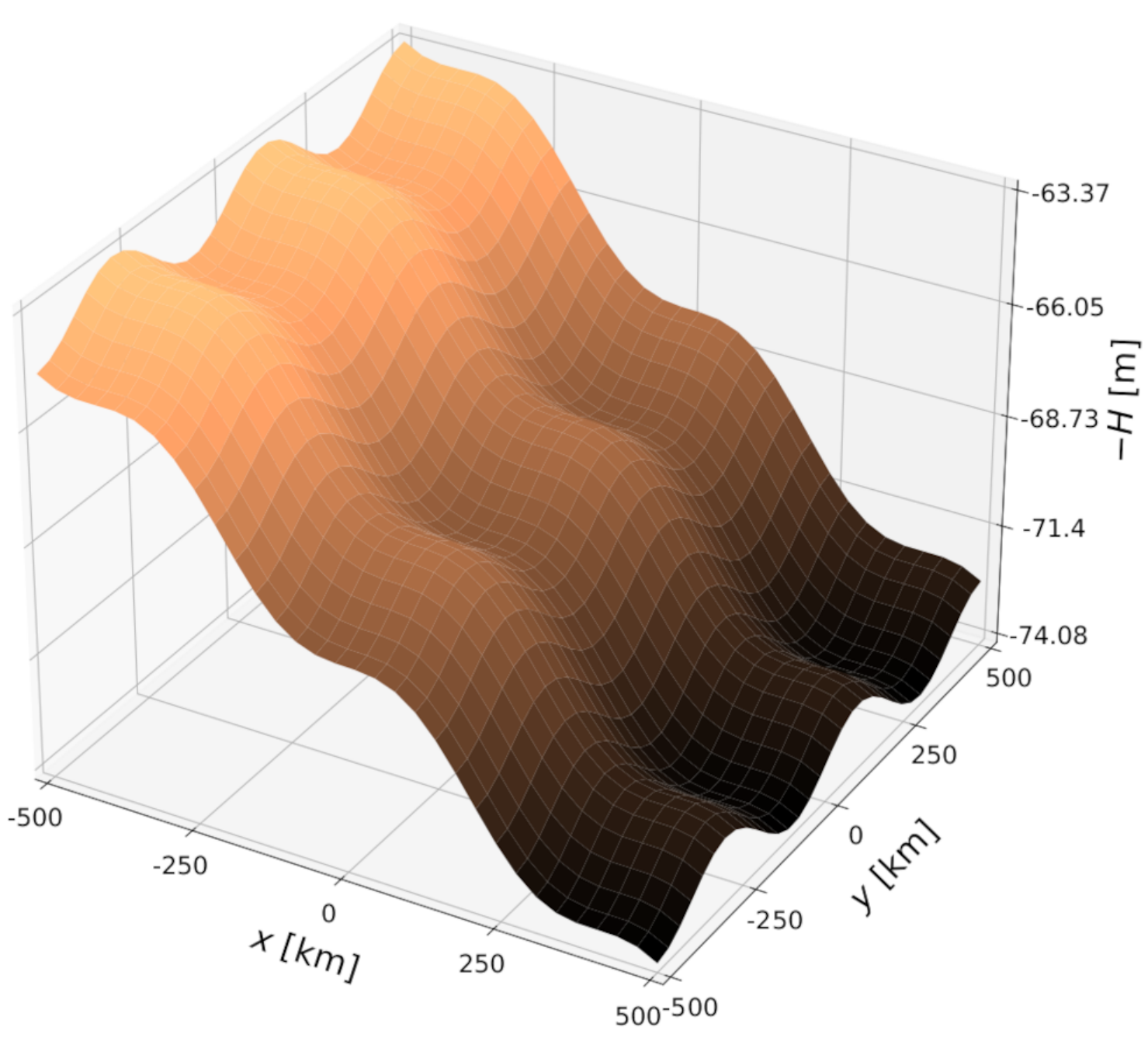}}
    \caption{Topography of the training and inference set. $H$ is located over a $1000km^2$ area. The training topography (left) is smooth, has a large depth range, and has different rotations and scales for each sequence in order to create a large variety in data. The inference topography (right) is more non-linear, bumpy, and is not rotated. It is the only topography for all sequences as it needs to be inferred by the models. The depth scale for the particular topography was randomly chosen as $\beta = 0.68$. It is the same in the left plot for a meaningful comparison.}
    \label{fig:data_topography}
\end{figure}

Training and inference processes follow the same scheme in each model and are worth explaining as they are different than in previous studies. In \cite{Horuz2022} and \cite{Horuz2023}, it was shown that a multi-BC training scheme improves performance at inference. In these previous studies, models received several sequences simultaneously in training (mini-batch learning scheme), each with a different set of boundary conditions. This allowed models to learn the effect of different BCs such that the trained model could infer the unknown BCs more accurately. With this insight, we trained models in multi-H scheme. Different sequences (in total $512$) each with a different topography were fed into the model. The ground-truth topography in data was calculated by the inverse of the tangent function such that the depth increases with distance and it has a smooth surface (see left plot in \autoref{fig:data_topography}). It is situated on a $1000 km^2$ structured mesh with an average depth of $100$ meters. Moreover, $H$ is rotated with a random angle $\phi \in [0, 2\pi]$ and the depth of $H$ is scaled with a random number $\beta \in [0.5, 1.0]$ such that the average depth is between $50-100m$ in each sequence. Additionally, the initial conditions, represented by a starting location of the Gaussian bump, are also randomly chosen. This vast diversity in topography data, with $H$ explicitly provided during training, is significantly beneficial for the models, exposing them to a wide range of topographies.

After training on various sequences and topographies, the task is to \textit{infer} a particular topography from data. However, accomplishing this task with a single sequence could cause problems because there may be artifacts in the data. Especially the Gaussian initialization could interfere the process and cause a gradient-induced peak in the topography. Consequently, relying on limited data could inadvertently result in models overfitting to these artifacts. Therefore different sequences with the same topography were generated (in total $256$). Topography used in inference data is bumpy, and more non-linear but has an average depth inside of the training depth range (see right plot in \autoref{fig:data_topography}). Identical to the training data, the topography was scaled with a randomly sampled number from $[0.5, 1]$ (the scale factor $\beta = 0.68$). Note that as the same topography is used for each inference sequence, they are also scaled by the same scale factor.

Similar to the previous experiments in \cite{Horuz2022} and \cite{Horuz2023}, $H$ is set as learnable parameter. Hence the number of total parameters is the same as the number of grid cells (in this study $32 \times 32$ grid). Given the gradient-based optimization framework and the use of mean squared error for loss calculation, this approach does not inherently consider the spatial relationships between grid cells of H. In preliminary studies, we realized that this approach leads to an unrealistic, distorted topography reconstruction. It was also not possible to apply active tuning \cite{ActiveTuning_Otte2020} as the topography does not change over time. To address this, a smoothness-constraint was introduced, penalizing significant deviations between adjacent cells to enhance the smoothness of surface. The regularization constant $\lambda$ is set to a small value (i.e. $5 \cdot 10^{-3}$ for PhyDNet and $5 \cdot 10^{-7}$ for the others) such that it was still possible to represent the non-linearity of the topography.

Another problem we faced in preliminary studies was the reconstruction of the edges. As stated before, $H$ is coupled with the velocity tensors and these have zero velocity on the edges due to the no-slip condition. As a result, the values of neighboring cells showed significant differences at the edges, leading to a deformed reconstruction. Another constraint was applied explicitly on the edges such that the values on the edges move closer to the ones towards the inside ($\lambda_{edge} = 5 \cdot 10^{-7}$). It is an unconventional employment of loss constraint but it basically regularizes edges especially because of their specific situation. The use of the $\lambda_{edge}$ regularization resulted in performance improvements across all models. However, both PhyDNet and DISTANA still struggled to produce accurate reconstructions along the edges. Results with and without edge reconstruction are provided for a detailed comparison in the following section.
\section{Experiments}
\label{sec:experiment}
\begin{table}[b!]
    \caption{Table presenting a comparison of various metrics including the number of parameters, training error, inference error, test error, and two types of reconstruction errors for each model. Inference error represents the minimum error achieved during model inference, while test error denotes the error of predictions while using the inferred topography. Additionally, two types of reconstruction errors are provided as the rooted mean squared error between the inferred and data topography. The inner reconstruction error excludes boundary errors and is computed on a smaller grid size of $28\times28$ instead of the original $32\times32$. Both training and test errors are averaged over batches for each epoch. The experiments were conducted $10$ times for each model to ensure statistical robustness.}
    \label{tab:top_rec}
    \centering
    \begin{tabularx}{320pt}{lccc}
        \toprule
        &
        \textbf{DISTANA} &
        \textbf{PhyDNet} &
        \textbf{FINN} \\
        \midrule
        \footnotesize{\# params} &
        $19,312$ &
        $185,444$ &
        $230$ \\
        \footnotesize{Train error} &
        $(6.4{\pm}1.6){\times}10^{-5}$ &
        $(5.2{\pm}2.0){\times}10^{-5}$ &
        $1{\times}10^{-5}\pm6{\times}10^{-9}$ \\
        \footnotesize{Infer. error} &
        $(7.8{\pm}1.8){\times}10^{-5}$ &
        $(1.6{\pm}1.1){\times}10^{-4}$ &
        $(2.2{\pm}0.1){\times}10^{-6}$ \\
        \footnotesize{Test error} &
        $(5.0{\pm}1.1){\times}10^{-5}$ &
        $(5.9{\pm}4.4){\times}10^{-5}$ &
        $(2.7{\pm}0.1){\times}10^{-6}$ \\
        \footnotesize{Full rec. error} &
        $0.913\pm0.125$ &
        $10.71\pm5.906$ &
        $0.262\pm0.013$ \\
        \footnotesize{Inner rec. error} &
        $0.516\pm0.121$ &
        $10.33\pm6.160$ &
        $0.237\pm0.018$ \\
        \bottomrule
    \end{tabularx}
\end{table}
\begin{figure}[t!]
    \centering
    \subfigure[DISTANA]{\includegraphics[width=0.48\textwidth]
          {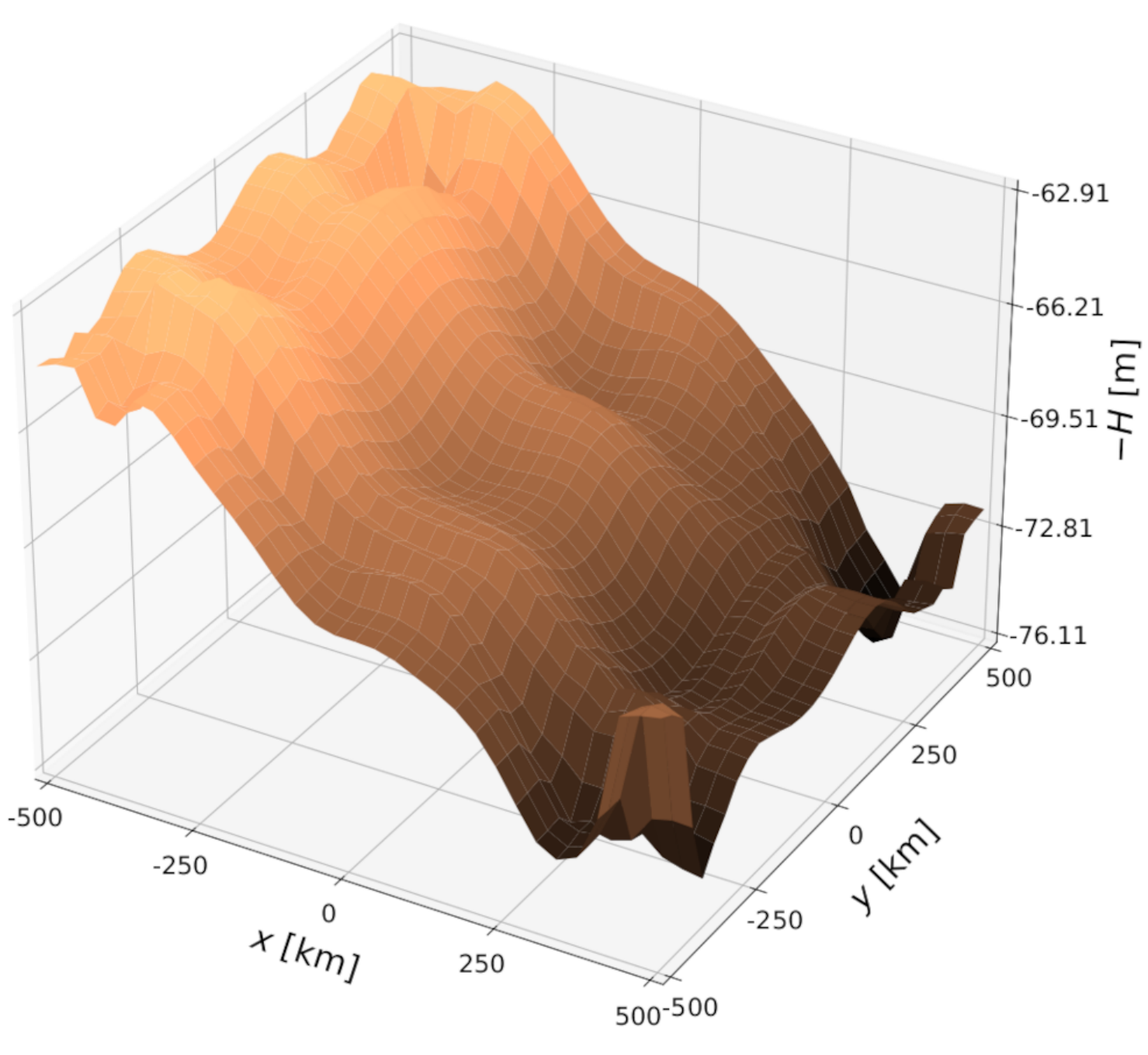}}
    \subfigure[PhyDNet]{\includegraphics[width=0.48\textwidth]
          {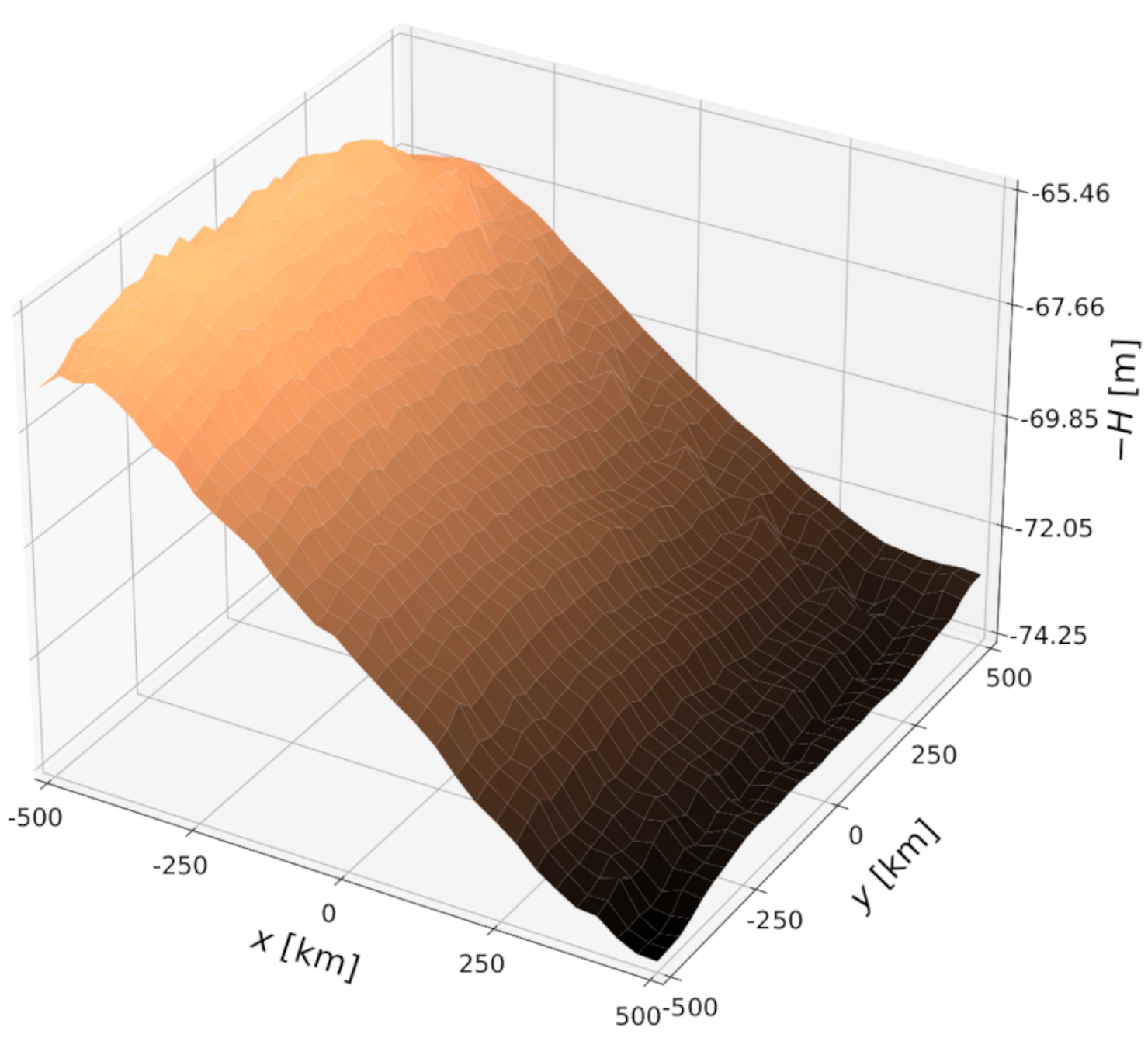}}
    \subfigure[FINN]{\includegraphics[width=0.48\textwidth]
          {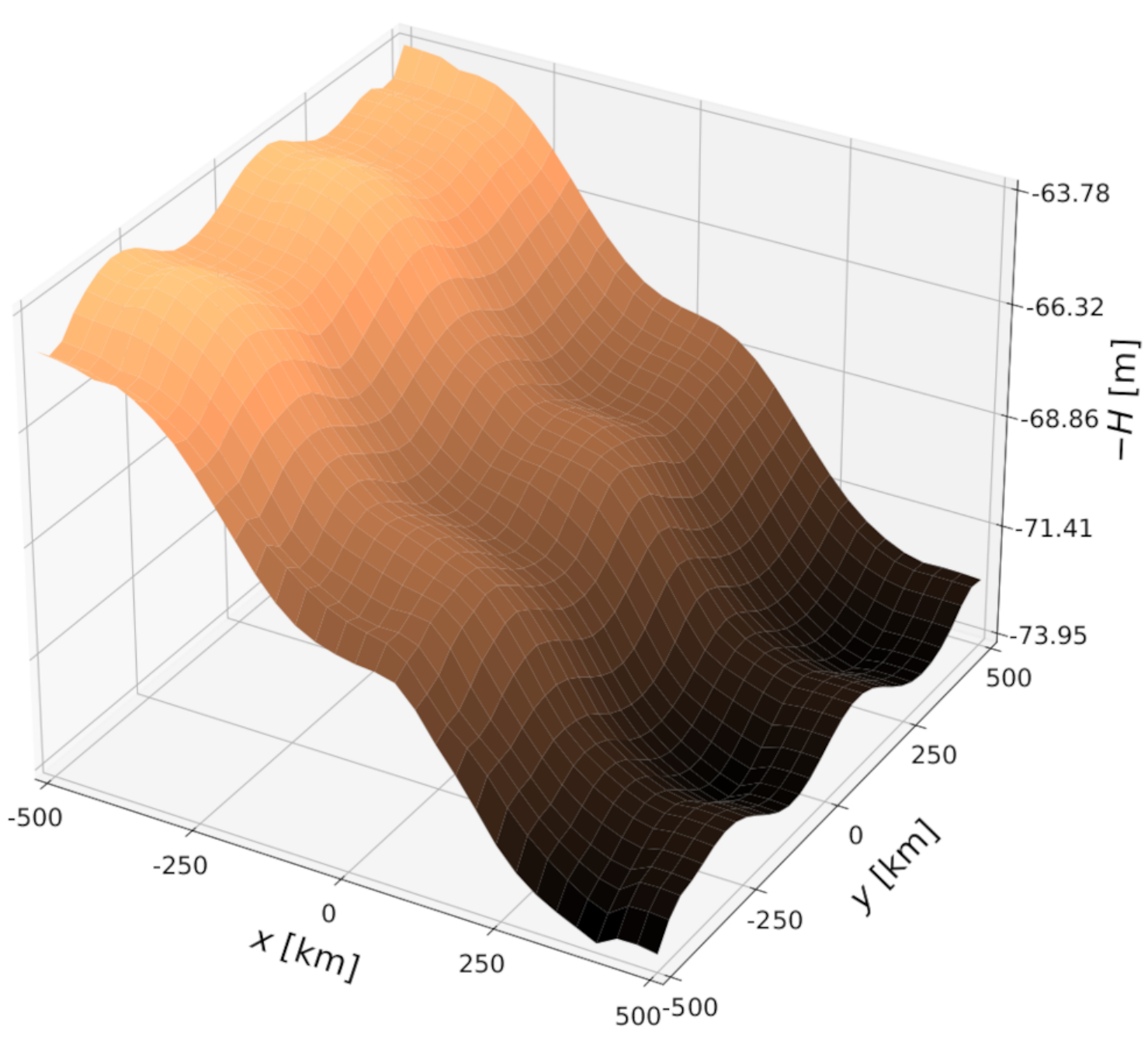}}
      \subfigure[Data]{\includegraphics[width=0.48\textwidth]{Figures/data_inference.pdf}}
    \caption{Different topographies over a $1000km^2$ area with varying depths. $-H$ is used for plotting to make the visualizations more realistic because physically higher depth means a bigger distance from the free surface. FINN's $H$ inference progress is given in \autoref{appendix:swe_top_inf}}
    \label{fig:top_3d}
\end{figure}
\begin{figure}[t!]
    \centering
    \subfigure[DISTANA]{\includegraphics[width=0.48\textwidth]
          {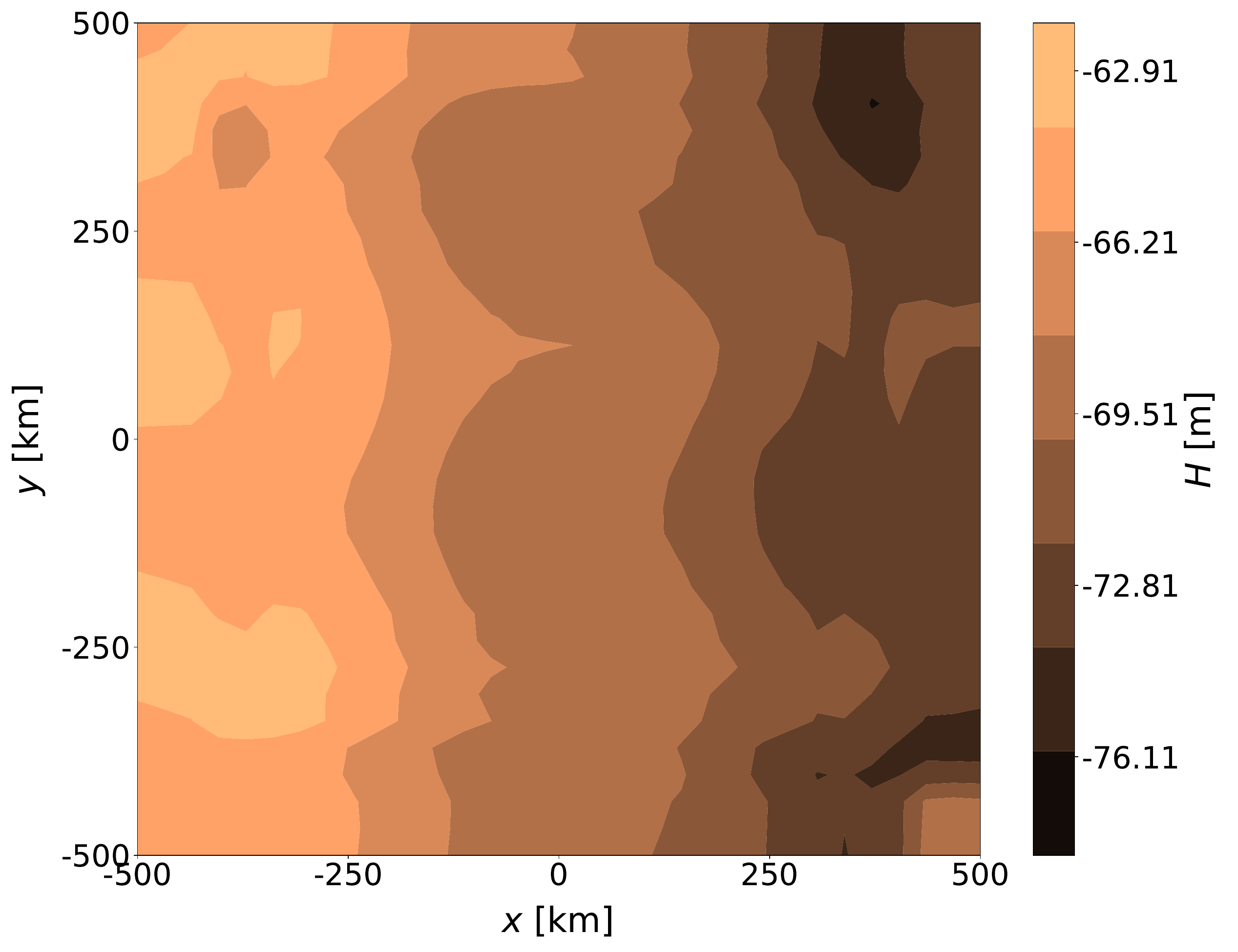}}
    \subfigure[PhyDNet]{\includegraphics[width=0.48\textwidth]
          {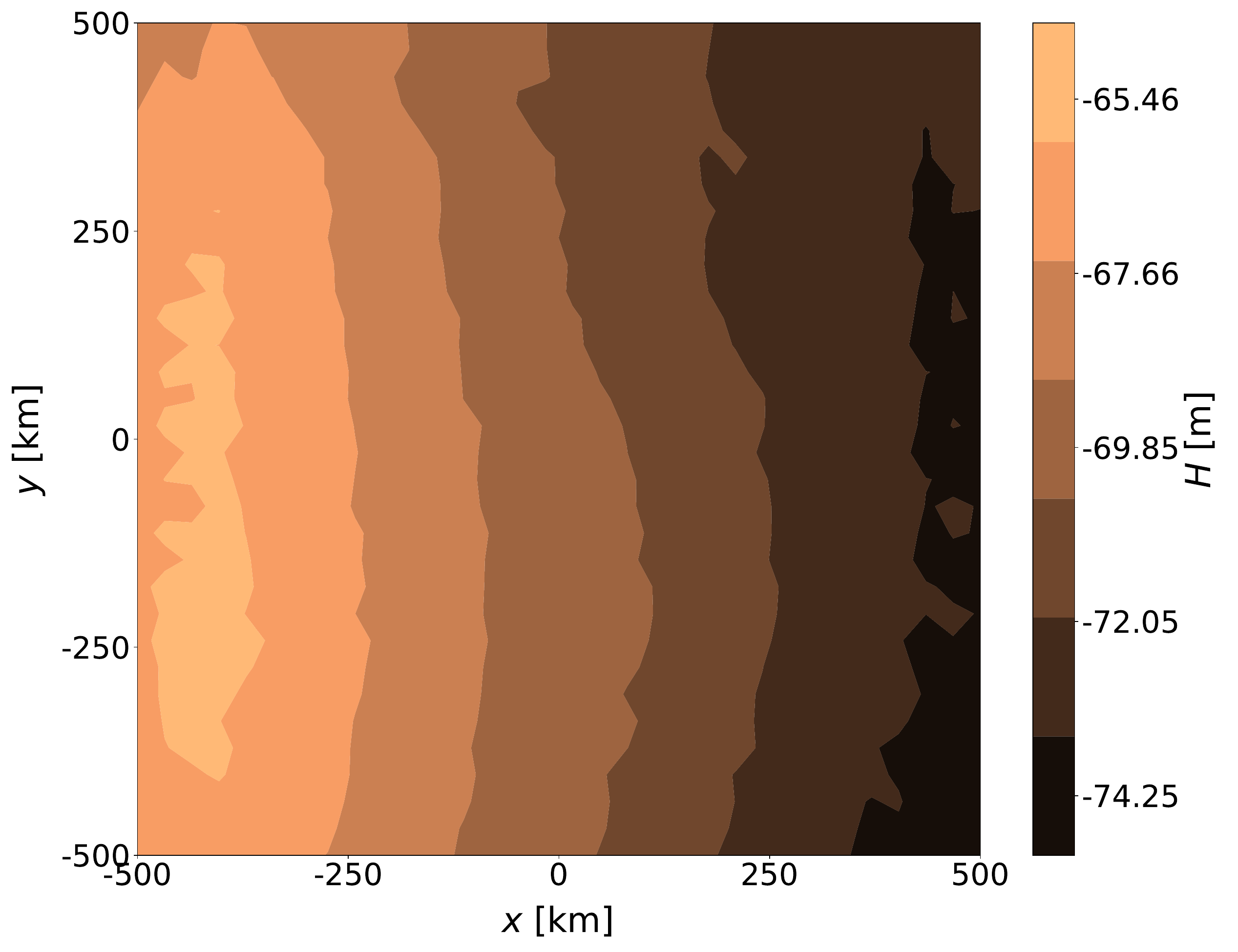}}
    \subfigure[FINN]{\includegraphics[width=0.48\textwidth]
          {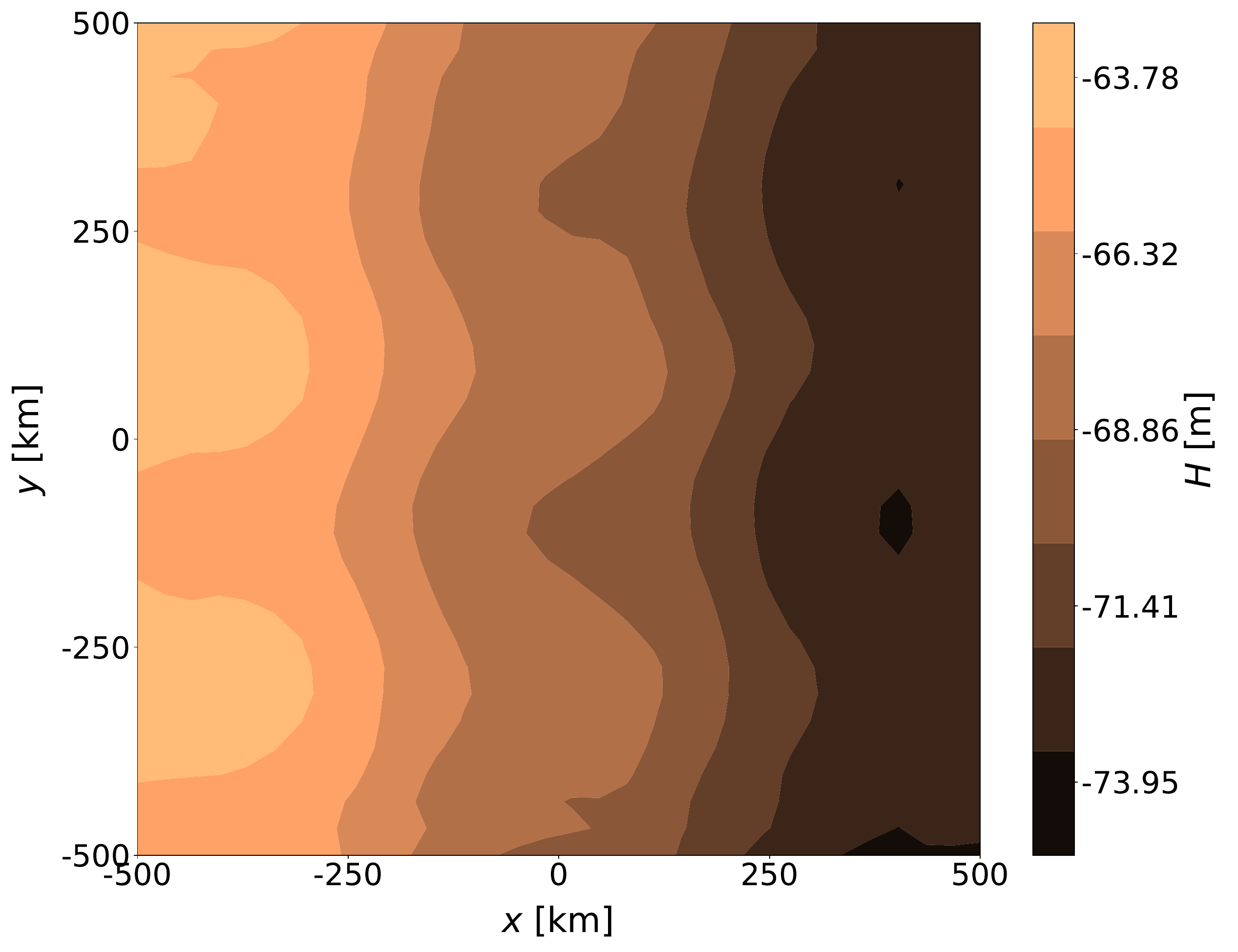}}
    \subfigure[Data]{\includegraphics[width=0.48\textwidth]
          {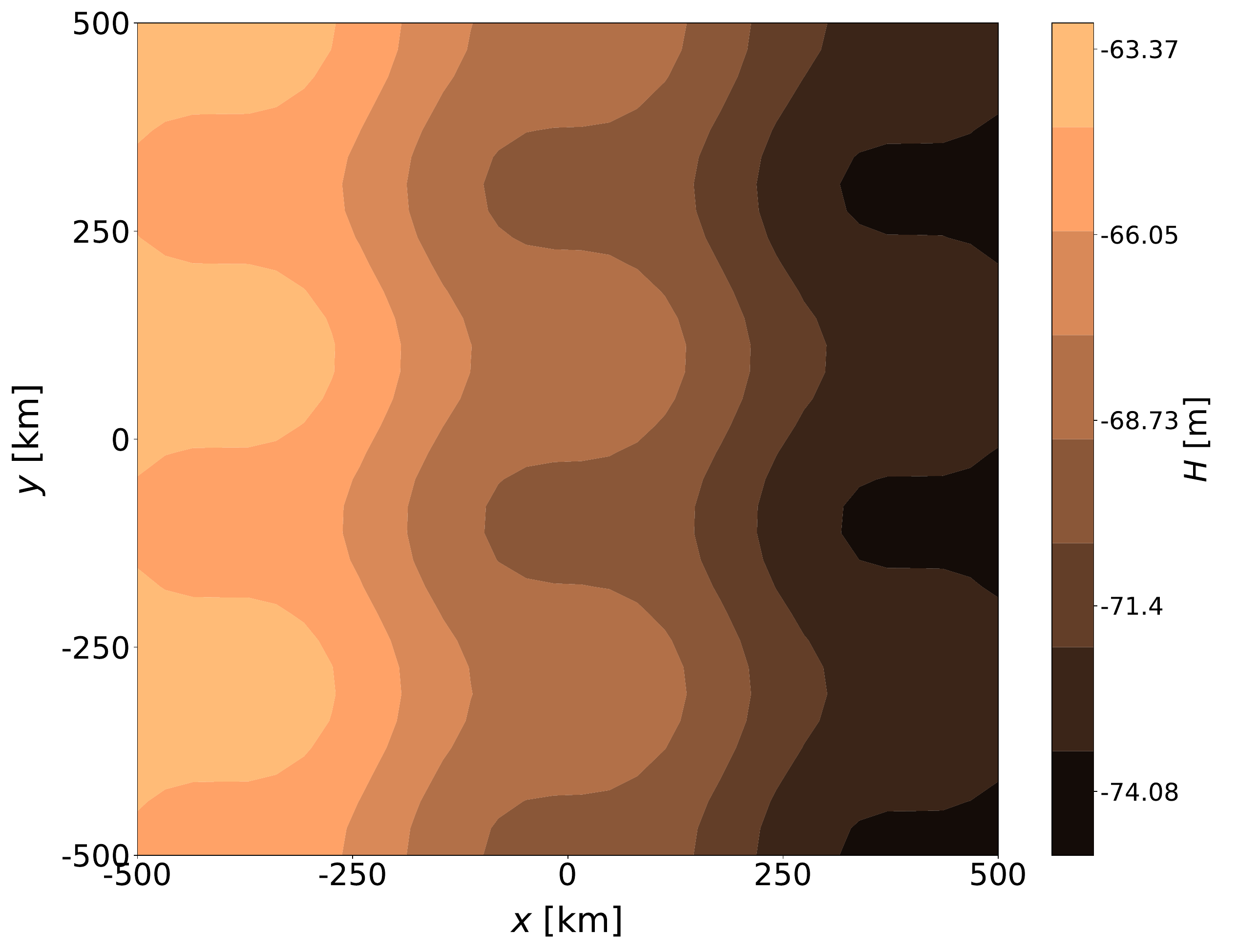}}
\caption{Data and the prediction shown as a contour plot. Again $-H$ is used for plotting to be aligned with the plots in \autoref{fig:top_3d}.}
\label{fig:top_contour}
\end{figure}
Throughout the experiments, the primary objective is to achieve optimal reconstruction results for each model. The findings are reported in \autoref{tab:top_rec}. It is noteworthy that DISTANA demonstrated performance comparable to that of FINN. Notably, given the wide range of values observed within the inference topography, falling within the interval $[63, 74]$, the errors attained by these two models carry particular significance. All three models successfully learned SWE and accurately predicted subsequent wave states with varying levels of accuracy. However, it is important to acknowledge the computational demands associated with PhyDNet, which imposes a significant computational burden. The slightly higher inference error exhibited by PhyDNet can be attributed to the application of a stricter grid regularizer aimed at enhancing reconstruction quality. Using the same $\lambda$ value for PhyDNet as other two modules has tripled the reconstruction error and the inferred topographies were predominantly characterized by spikes.

The most accurate topography reconstructions for each model are depicted in \autoref{fig:top_3d} and \autoref{fig:top_contour}. While \autoref{fig:top_3d} offers a realistic and intuitive portrayal of the topographies, \autoref{fig:top_contour} provides a clearer visualization of discrepancies. The topography inference process of FINN is also given in much detail in \autoref{appendix:swe_top_inf}. These visualizations confirm that FINN and DISTANA manage to reconstruct the topography although FINN's precision is superior. In particular, DISTANA exhibits notable deviations along the edges, attributable to inadequacies in boundary condition implementation. As it was also stated in the previous works \cite{karlbauer2021composing,Horuz2023}, FINN rigorously applies boundary conditions based on the governing equations. However, DISTANA and PhyDNet rely on convolution operations for this purpose, leading to poorer edge reconstruction. This distinction underscores FINN's advantage in handling boundary conditions. Therefore, to ensure a fair comparison, we show two different reconstruction errors in \autoref{tab:top_rec}. \emph{Full error} encompasses the error of the entire spatial domain, i.e. the grid size is $32\times32$. \emph{Inner domain error}, on the other hand, omits $5$ data points on each boundary for a fair comparison, i.e. the grid size is $28\time28$. Indeed, the error difference between DISTANA and FINN is closer within the inner domain, however, DISTANA's error remains approximately double of FINN's error. PhyDNet, on the contrary, fails to produce meaningful reconstruction. The high standard deviation of PhyDNet's prediction is also another argument that it does not reliably perform on this task.

One important point that could catch the eyes of the reader is the difference between training and test error by FINN. Normally it is expected to have smaller training error. Nevertheless, the conventional error computation is not used in this experiment due to the distinct topographies present in the training and testing datasets. Consequently, this leads to different error regimes. Although the topography of the training data is smooth and mildly non-linear, it spans from $55$ to $82$ with a range of $82 - 55 = 27$. This range is $2.45$ times bigger than the topography values of the inference/test set which has a range of $74 - 63 = 11$. Thus, the training set has a larger error regime. This case has been verified numerically as well. We have observed that a model trained without inference on both datasets exhibited an error scheme similar to the results in \autoref{tab:top_rec}, confirming the impact of topography on the training process.
\section{Discussion}
\label{sec:discussion}
The aim of this study was to assess the capabilities of DISTANA, PhyDNet, and FINN in inferring the underwater topography depending on the shallow water equations. Compared with the previous investigations\cite{praditia2021finite,praditia2022learning,karlbauer2021composing,Horuz2022,Horuz2023}, SWE differ in two aspects. Firstly, behaviors of SWE are characterized not by a single PDE but by a system of three coupled PDEs. Secondly, SWE is a dynamic equation such that the waves move until the end of the sequence. All the previously examined systems by FINN reach a sort of equilibrium state after a while. To adapt FINN to these novelties, it was modified in different ways which are explained in \autoref{sec:swe} and \autoref{sec:und_top_rec}. These developments on FINN are not only useful in reaching the goal of this study; they also pave the way for a broader application of the model across various domains. 

FINN produced remarkable results. Along with the low test error, it produced accurate reconstruction of the underwater topography (see Figures \ref{fig:top_3d} and \ref{fig:top_contour}). Although inferior to FINN, DISTANA managed to successfully solve the SWE system and inferred the underlying topographical structure. This aligns findings from the previous study by \cite{karlbauer2021latent}, which demonstrated DISTANA's capabilities in inferring latent land-sea mask. PhyDNet, despite its physics-aware concept, managed to model the SWE but struggled to accurately reconstruct a topography. PhyDNet is a large model, compared to the other models in this study. This causes an overfitting scenario to the training data such that the model is too complex for the problem and cannot give enough \textit{importance} to the embedded topography.
\section{Conclusion}
\label{sec:conclusion}
In addressing physical problems, pure ML models often stay too \textit{universal} especially in inferring unknown structures in data. This study underscores the crucial need for a physically structured model, encapsulating application-specific inductive bias, to complement the learning abilities of neural networks. FINN encompasses these two aspects by implementing multiple MLP modules through mathematical composition to impose physical constraints. This architecture allows FINN to establish unknown structures in data at inference, setting it apart from other physics-aware ML models. Wide range of spatiotemporal problems can be characterized by  advection-diffusion processes and FINN can solve them. In addition to that, FINN can also generalize to certain unknown structures in data with high precision. The current study has demonstrated the potential of FINN to accurately infer and generalize unknown data structures representing the underwater topography. However,  the potential applications and scope for FINN-based architectures stretch far beyond current achievements, revealing a vast directions for possible research. The long-term aim could be the application of FINN to a variety of real-world scenarios of a larger scale as an extension to the previous work by \cite{praditia2022learning}. Especially considering the SWE as highly relevant in modeling wave dynamics in coastal regions including tsunamis \cite{tsunami2005}. The prospect of FINN reconstructing coastal underwater topography of coastal regions from real-world data presents a relevant pathway for exploration. We do hope, that this article opens a road towards real-world data inference. Furthermore, given that SWE represents a simplified version of Navier-Stokes equations, further research efforts could be directed towards integrating more complex versions of these equations into FINN's framework. This advancement will significantly enhance FINN's utility and broaden its application in accurately understanding and predicting natural phenomena.

\bibliographystyle{plain}
\bibliography{2024-FINNInference}
\newpage
\appendix

\section{Derivation of Shallow Water Equations}\label{appendix:swe_derivation}
Here, we derive the shallow water equations from scratch. As shallow water equations deal with incompressible fluids (e.g. water), the equation $\nabla \cdot \mathbf{v} = 0$ will be the equation from which we will derive parts of SWE.

\subsection{Continuity Equation in SWE}
SWEs are PDEs governing fluids in coastal regions, rivers and channels. They are different forms of continuity and momentum equations. The mass continuity and momentum equations are ubiquitous in fluid mechanics and also the governing equations in shallow water equations.

The shallow water assumption will make the difference in the derivation of momentum and continuity equations and it is defined as follows:
\begin{equation}\label{eq:sw}
    \frac{H}{\lambda} \ll 1
\end{equation}
where $H$ is the mean depth and $\lambda$ is the wavelength. Before starting the derivation, we can look at what shallow water assumption effectively means. As we assume that the mean depth $H$ is much smaller than the wavelength $\lambda$, the vertical velocities are much smaller than the horizontal velocity. This realization helps us to make important conclusions subject to the definition of pressure. Pressure of fluids is defined as follows:
\begin{equation}
    p = \rho g h
\end{equation}
where $p$ is pressure, $\rho$ corresponds to the density of the fluid, $g$ is equal to the acceleration through gravity and $h$ is the depth of the fluid. This equation can be applied to the shallow water assumption at a variable height $z$ (c.f. \autoref{fig:swe}):
\begin{equation}\label{eq:p}
    p = \rho g(H + \eta(x,y,t) - z)
\end{equation}
Note that as $z$ increases, $p$ decreases. This is aligned with the general knowledge that the pressure increases with the depth in water. The horizontal pressure gradients derived from \autoref{eq:p} are as follows:
\begin{equation}\label{eq:p2}
    \frac{\partial p}{\partial x} = \rho g \frac{\partial \eta}{\partial x} \hspace{0.5cm}\text{and}\hspace{0.5cm} \frac{\partial p}{\partial y} = \rho g \frac{\partial \eta}{\partial y}
\end{equation}
This means that the horizontal motion is depth-independent. Therefore, the horizontal velocities ($u$ and $v$ for $x$ and $y$ directions, respectively) do not depend on $z$. The depth-independence of the horizontal velocities is an outcome of the shallow water assumption and will be influential in the derivation.

Another important concept is \textit{no-slip condition}. In a nutshell, it states that the velocity of a fluid on a solid surface is zero. Accordingly the bottom and side boundary conditions will be zero such that $u(\Omega, t) = v(\Omega, t) = 0$ where $\Omega$ corresponds to the boundaries except the free surface. Nonetheless the free surface boundary condition is still need to be applied. We will see how it is applied later. First we will derive the continuity equation. Let $\mathbf{v} = [u, v, w]$ be the vector of velocity functions. Then
\begin{align}
    0 &= \nabla\cdot\mathbf{v} \label{eq:c1}\\
    \int_{0}^{H+\eta}0 &= \int_{0}^{H+\eta} \left[\frac{\partial u}{\partial x} + \frac{\partial v}{\partial y} + \frac{\partial w}{\partial z}\right]dz \label{eq:c2}\\[1em] 0 &= \int_{0}^{H+\eta}\frac{\partial u}{\partial x}dz + \int_{0}^{H+\eta}\frac{\partial v}{\partial y}dz + \int_{0}^{H+\eta}\frac{\partial w}{\partial z}dz \label{eq:c3}\\[1em]
    &= \frac{\partial u}{\partial x}\int_{0}^{H+\eta}1dz + \frac{\partial v}{\partial y}\int_{0}^{H+\eta}1dz + w\vert_{z=0}^{z=(H+\eta)} \label{eq:c4}\\[1em]
    &= (H+\eta)\frac{\partial u}{\partial x} + (H+\eta)\frac{\partial v}{\partial y} + w(H+\eta) - \red{w(0)} \label{eq:c5}
\end{align}
\autoref{eq:c1} is basically the definition of divergence for incompressible fluids. In \autoref{eq:c2} divergence is written explicitly and the integral is taken on both sides. As $u$ and $v$ do not depend on $z$ (shallow water assumption), they can be considered as constants and taken out of the integral. This is done in \autoref{eq:c4}. Furthermore, integral of a derivative of a function is the function itself (fundamental theorem of calculus). Therefore $w$ is evaluated at $z=(H+\eta)$ and $z=0$ in \autoref{eq:c4}. Lastly, because of the no-slip condition $w(0) = 0$.

As stated above, free surface boundary condition should be applied on the continuity equation. In order to do that, we need to introduce the Euler equation for total (also called material) derivative. It is a fairly common equation in many different fields of physics and probably every physics bachelor student is acquainted with total derivatives. Interested readers are referred to lecture $1$ in \cite{segur2009} or chapter $5$ in \cite{tritton2007}. The material derivative is defined as:
\begin{align}\label{eq:euler}
    \frac{D\phi}{Dt} &= \frac{\partial \phi}{\partial t} + \mathbf{v} \cdot \nabla \phi \nonumber \\
    &= \frac{\partial \phi}{\partial t} + u\frac{\partial \phi}{\partial x} + v\frac{\partial \phi}{\partial y} + w\frac{\partial \phi}{\partial z}
\end{align}
where $\phi$ is a function of $(x, y, z, t)$. \autoref{eq:euler} means that the total derivative of a function with respect to time gives the rate of change of that particular function over time. In other words, the total derivative denotes the rate of change of \textit{any} quantity on which it operates. It is a general formulation for differentiation in fluid mechanics.

We will use \autoref{eq:euler} to find the value at the free surface as it will be the boundary condition. Accordingly, if we take the total derivative of $(H+\eta)$, we can find the rate of vertical change (motion at the free surface) which should be equal to the boundary condition. \autoref{eq:euler} can be written for $(H+\eta)$ as follows:
\begin{equation}\label{eq:bc}
    \frac{D(H+\eta)}{Dt} = \frac{\partial (H+\eta)}{\partial t} + u\frac{\partial (H+\eta)}{\partial x} + v\frac{\partial (H+\eta)}{\partial y} + w\red{\frac{\partial (H+\eta)}{\partial z}}
\end{equation}
$\eta$ is a function of $x$, $y$ and $t$. Thus it does not depend on $z$. The partial derivative of $\eta$ with respect to $z$ (depicted in red) should be $0$. Remember that \autoref{eq:bc} gives us the boundary condition. It is also possible to set the boundary condition explicitly. Intuitively, it should be the vertical velocity at the free surface, namely $w(H+\eta)$. Therefore \autoref{eq:bc} should be equal to $w(H+\eta)$. Additionally note that $H$ is the mean depth of the fluid. It is a constant and do not change neither in time nor in space. Its derivative is therefore always $0$. The respective changes on \autoref{eq:bc} can be done as follows:
\begin{equation}
    w(H+\eta) = \frac{\partial \eta}{\partial t} + u\frac{\partial \eta}{\partial x} + v\frac{\partial \eta}{\partial y}
\end{equation}
Now we can substitute $w(H+\eta)$ in \autoref{eq:c5}:
\begin{equation}\label{eq:c6}
    \blue{(H+\eta)\frac{\partial u}{\partial x}} + \darkgreen{(H+\eta)\frac{\partial v}{\partial y}} + \frac{\partial \eta}{\partial t} + \blue{u\frac{\partial \eta}{\partial x}} + \darkgreen{v\frac{\partial \eta}{\partial y}}
\end{equation}
Lastly, we will use the chain rule to simplify \autoref{eq:c6}. Here is a formulation of the chain rule specific to our case:
\begin{align}
    \frac{\partial}{\partial x}[u(H+\eta)] &= u\frac{\partial(H+\eta)}{\partial x} + (H+\eta)\frac{\partial u}{\partial x} \nonumber \\
    &= u\frac{\partial\eta}{\partial x} + (H+\eta)\frac{\partial u}{\partial x}
\end{align}
Now we can apply the chain rule to the respective blue and green terms.
\begin{equation}\label{eq:cont}
    \frac{\partial \eta}{\partial t} + \frac{\partial}{\partial x}[u(H+\eta)] + \frac{\partial}{\partial y}[v(H+\eta)] = 0
\end{equation}
\autoref{eq:cont} is the mass continuity function with the shallow water assumption. It shows how the mass (particles) changes in time and space.

As it is stated before, shallow water equations are the different forms of continuity and momentum equations. We derived the continuity equation from the divergence theorem. The next step is to derive the momentum equations with the shallow water assumption.

\subsection{Momentum Equations in SWE}
Momentum and pressure is highly related in fluid dynamics as the pressure is a decisive force of the movement of the particles and part of this force comes from the particles' momentum change. Therefore, in this part, we will make use of the pressure gradients in \autoref{eq:p2} and also the total derivative in \autoref{eq:euler}. The Euler equation in its general form can represent the change of \textit{any} quantity. Thus it can be used to find the change of pressure in a particular direction. In that case change of pressure in $x$-direction would be equal to the total derivative of the velocity in that direction, i.e. $u(x,t)$. Another way to find the change of pressure in $x$-direction is basically taking the derivative of pressure, which is $-\partial p/\partial x$ by convention (see chapter $2.2$ in \cite{tritton2007} for further details). We can combine the two formulations as follows:
\begin{equation}
    -\frac{\partial p}{\partial x} = \frac{Du}{Dt} = \frac{\partial u}{\partial t} + u\frac{\partial u}{\partial x} + v\frac{\partial u}{\partial y} + w\frac{\partial u}{\partial z}
\end{equation}
We can replace $-\partial p/\partial x$ as in \autoref{eq:p2}. Furthermore, as we consider a constant density, we omit $\rho$ from the equation.
\begin{equation}
    -g\frac{\partial \eta}{\partial x} = \frac{\partial u}{\partial t} + u\frac{\partial u}{\partial x} + v\frac{\partial u}{\partial y} + w\frac{\partial u}{\partial z}
\end{equation}
Note that $u$ does not depend on $y$ or $z$. The respective derivatives should be $0$.
\begin{equation}\label{eq:m1}
    -g\frac{\partial \eta}{\partial x} = \frac{\partial u}{\partial t} + u\frac{\partial u}{\partial x}
\end{equation}
The same process can be applied to the velocity $v$ in $y$-direction analogously:
\begin{equation}\label{eq:m2}
    -g\frac{\partial \eta}{\partial y} = \frac{\partial v}{\partial t} + v\frac{\partial v}{\partial y}
\end{equation}
\autoref{eq:m1} and \autoref{eq:m2} are the momentum equations with the shallow water assumption which are derived from the equation depicting the change in pressure.

\begin{figure}[h!]
    \caption{\textit{Visualization of SWE.} The first plot is the initial state defined as a Gaussian bump in \autoref{eq:gauss}, which creates an effect of an enormous rock dropped into water. The plane represents an area with the size of $1000km^2$ whereas the vertical axis shows the waves in meters. Waves are reflected from the edges back into the domain. The time is shown on top of the figures and the simulation length spans $19.71$ hours in total.}
    \subfigure{\includegraphics[width=0.32\textwidth]{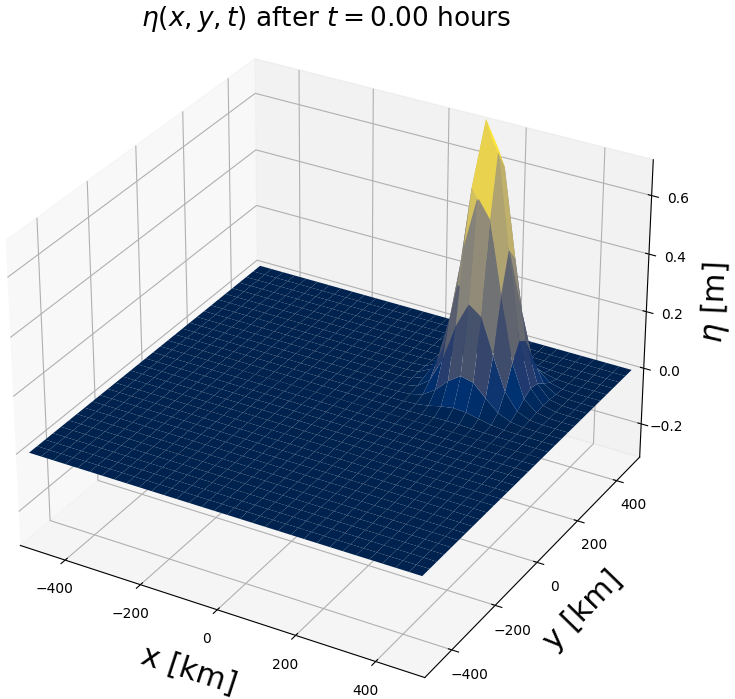}}
    \subfigure{\includegraphics[width=0.32\textwidth]{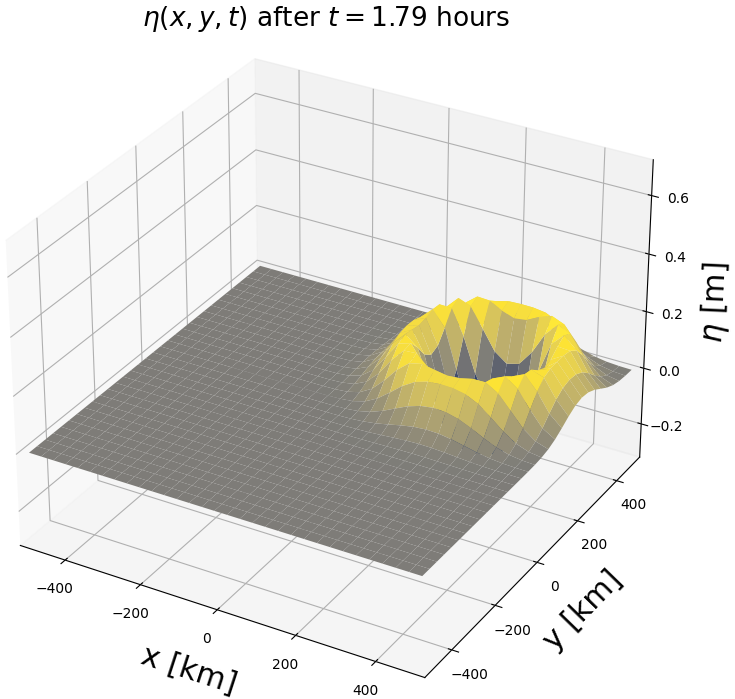}}
    \subfigure{\includegraphics[width=0.32\textwidth]{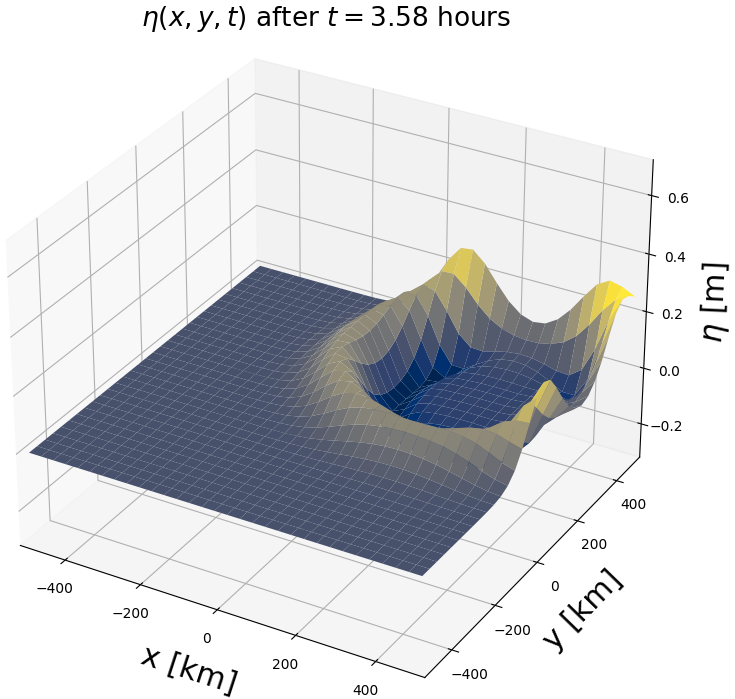}}\\
    \subfigure{\includegraphics[width=0.32\textwidth]{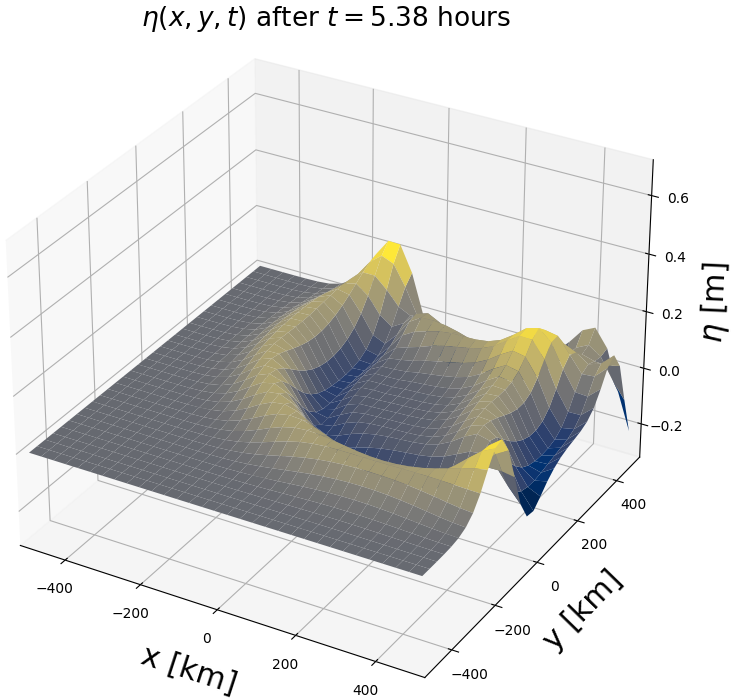}}
    \subfigure{\includegraphics[width=0.32\textwidth]{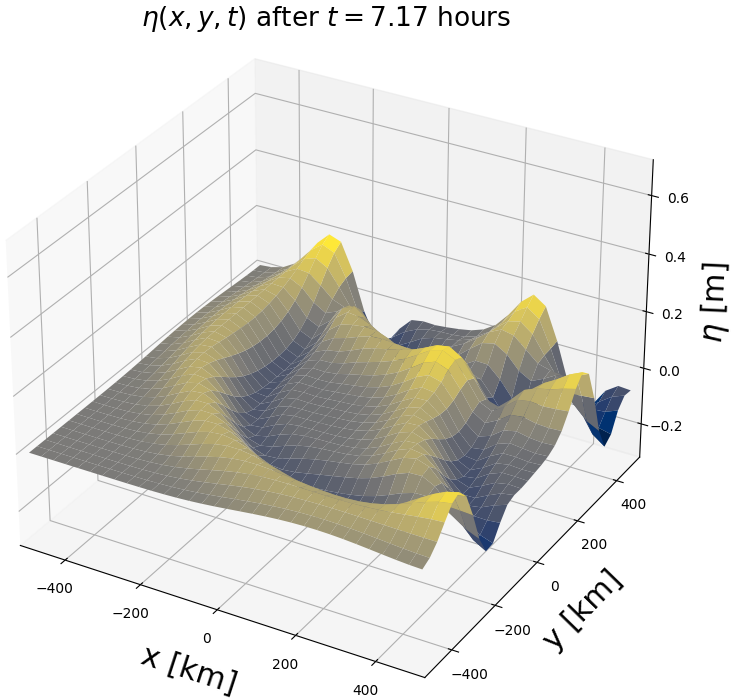}}
    \subfigure{\includegraphics[width=0.32\textwidth]{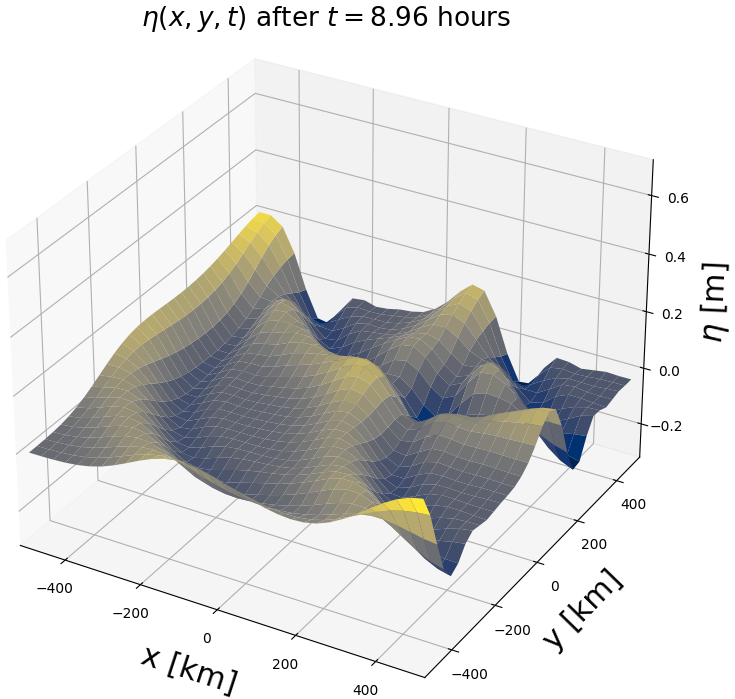}}\\
    \subfigure{\includegraphics[width=0.32\textwidth]{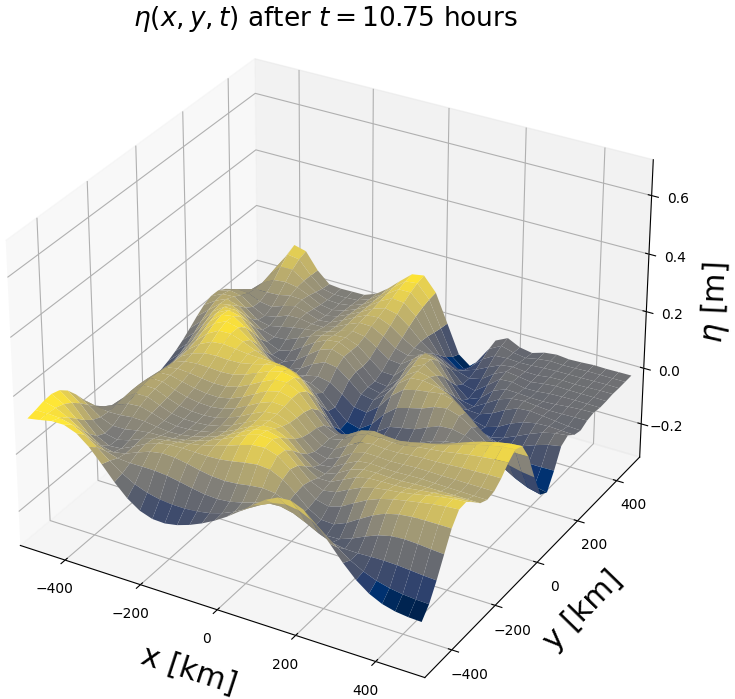}}
    \subfigure{\includegraphics[width=0.32\textwidth]{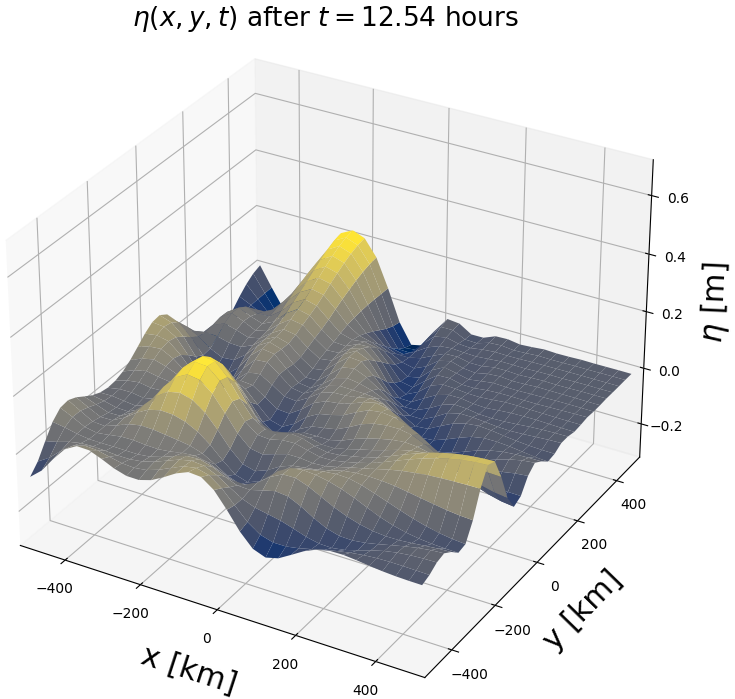}}
    \subfigure{\includegraphics[width=0.32\textwidth]{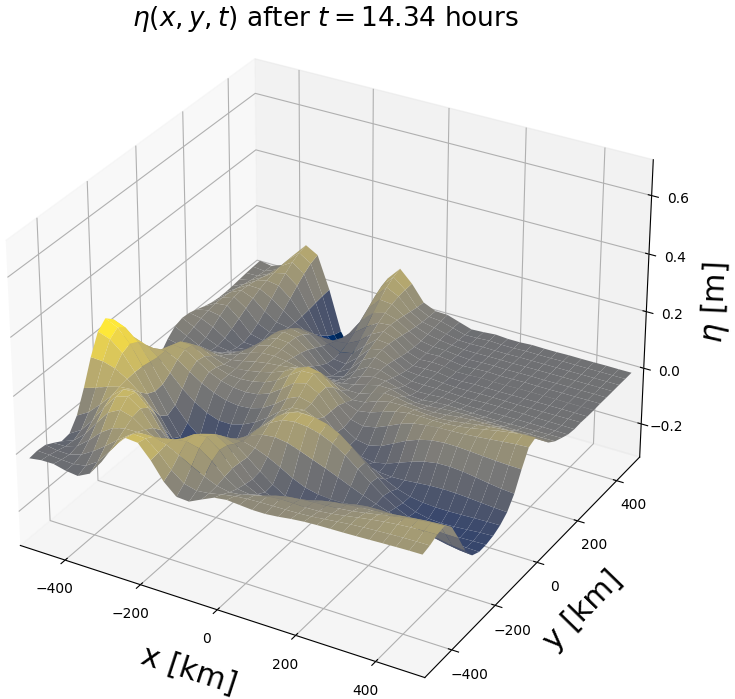}}\\
    \subfigure{\includegraphics[width=0.32\textwidth]{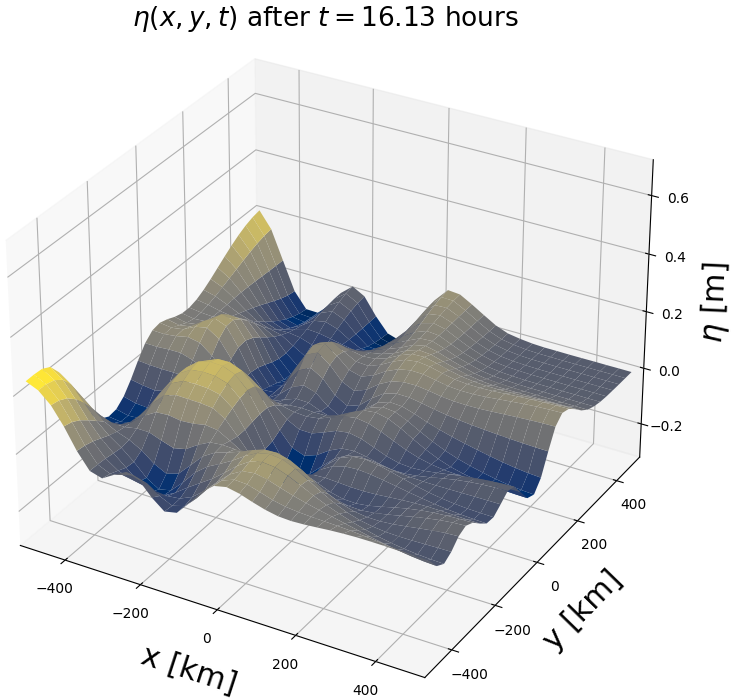}}
    \subfigure{\includegraphics[width=0.32\textwidth]{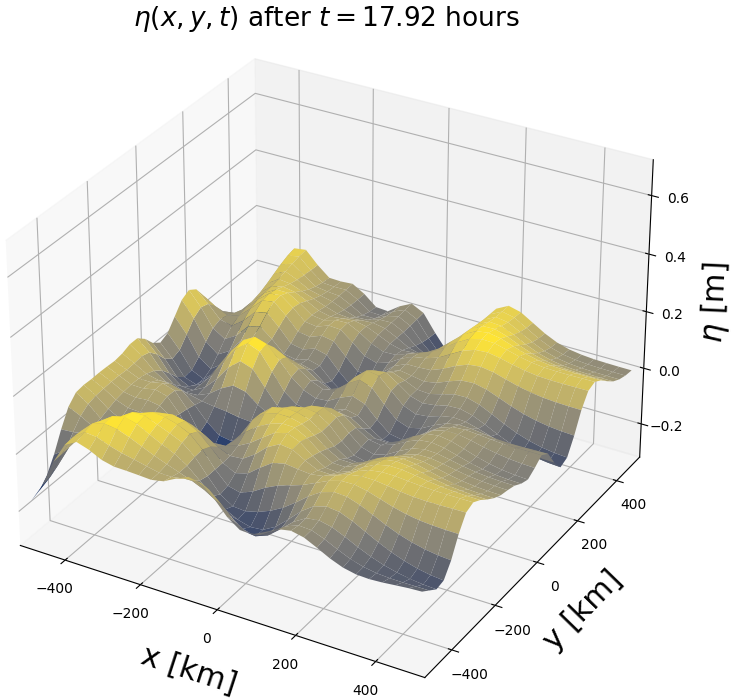}}
    \subfigure{\includegraphics[width=0.32\textwidth]{Figures/swe_eq/eta_100.png}}\\
    \label{fig:swe_vis}
\end{figure}

The momentum equations we derived are in non-linear form. However, we use linearized momentum equations in this work. As it will be clear in the following derivation, the reason for this linearization is because the non-linear terms has a small effect on the equation such that they are negligible. Thus we linearize the momentum equations. Accordingly, we need to consider small disturbances about the fluid at rest:
\begin{equation}
    u = 0 + \frac{\partial u}{\partial x}dx \hspace{0.3cm} \text{and} \hspace{0.3cm} v = 0 + \frac{\partial v}{\partial y}dy
\end{equation}
This assumption states the perturbations in the velocity have negligible amplitude and hence the first term in perturbation series (a concept similar to the Taylor expansion series) can be taken as zero. Now we can apply linearity to the momentum equation:
\begin{equation}
    -g\frac{\partial \eta}{\partial x} = \frac{\partial u}{\partial t} + \left(\frac{\partial u}{\partial x} \right)^{2}dx
\end{equation}
As we would like to derive the linearized momentum equations, we can neglect the second-order terms giving us the final version of the momentum equations as follows:
\begin{align}
    \frac{\partial u}{\partial t} &= -g\frac{\partial \eta}{\partial x} \label{eq:mu}\\ \frac{\partial v}{\partial t} &= -g\frac{\partial \eta}{\partial y} \label{eq:mv}
\end{align}
The resulting equations (\autoref{eq:cont}, \ref{eq:mu} and \ref{eq:mv}) represent the mass continuity equation in its non-linear form and linearized momentum conservation equations. These are used to simulate the dynamics of shallow water processes. \autoref{fig:swe_vis} shows a series of plots that are taken from the simulation of the SWE.
\newpage
\newpage
\section{Visualization of topography inference}
\label{appendix:swe_top_inf}
Following are the sampled images from the inference process of the underwater topography, i.e. $H$. The plots depict how the non-linear topography reconstructed by FINN from a flat surface in around $1400$ iterations. Note that inference process took in total $1600$ iterations. The fully reconstructed topography can be seen in \autoref{fig:top_3d}.
\begin{figure}[ht]
    \subfigure{\includegraphics[width=0.33\textwidth]{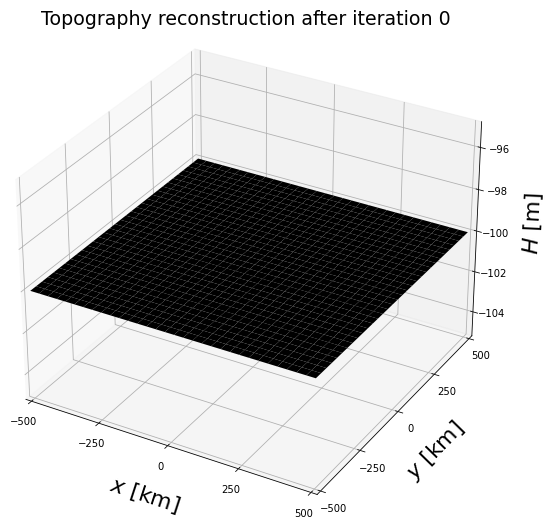}}
    \subfigure{\includegraphics[width=0.33\textwidth]{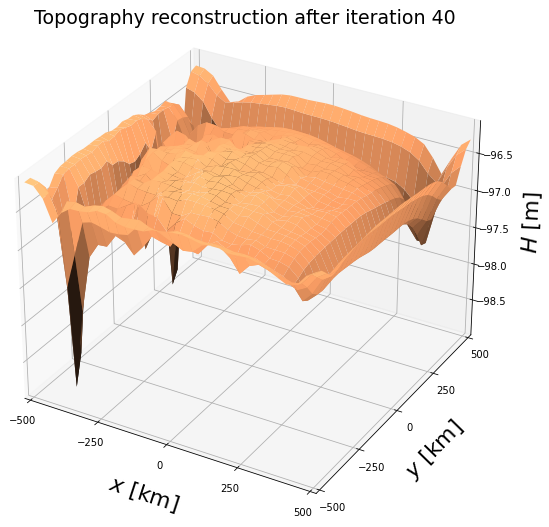}}
    \subfigure{\includegraphics[width=0.33\textwidth]{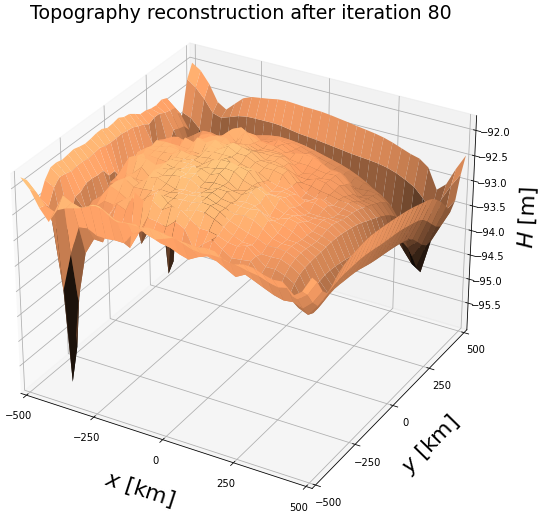}}\\
    \subfigure{\includegraphics[width=0.33\textwidth]{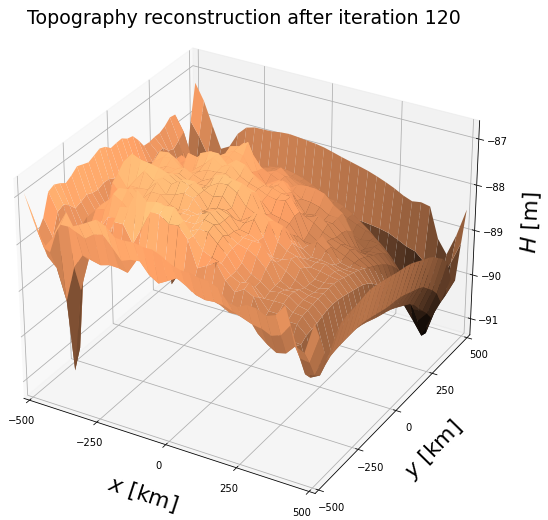}}
    \subfigure{\includegraphics[width=0.33\textwidth]{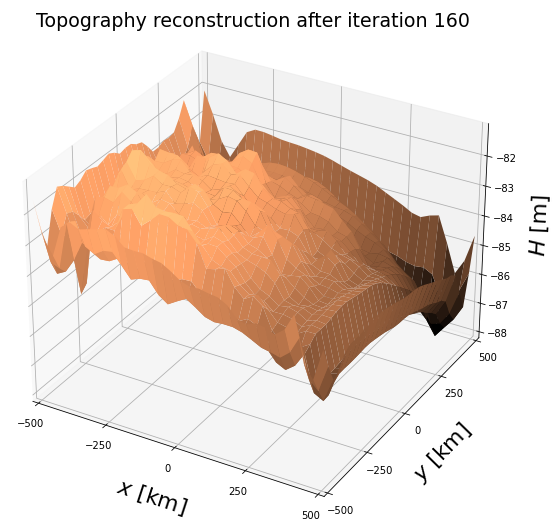}}
    \subfigure{\includegraphics[width=0.33\textwidth]{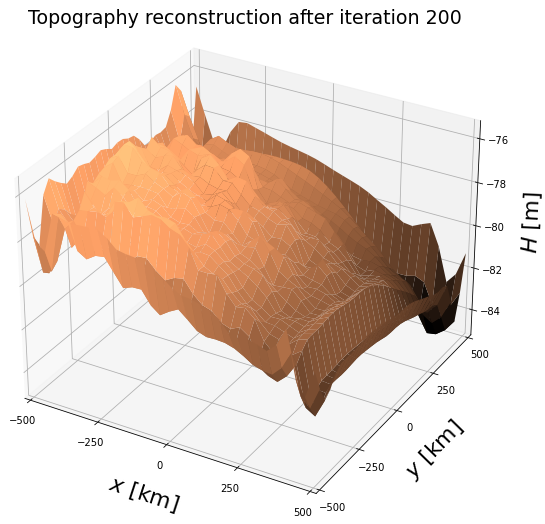}}\\
    \subfigure{\includegraphics[width=0.33\textwidth]{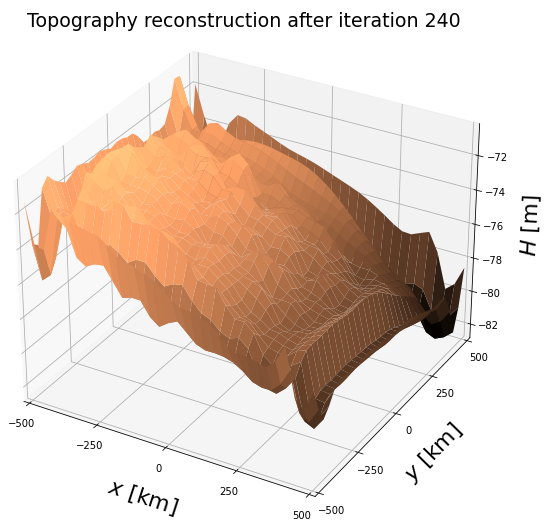}}
    \subfigure{\includegraphics[width=0.33\textwidth]{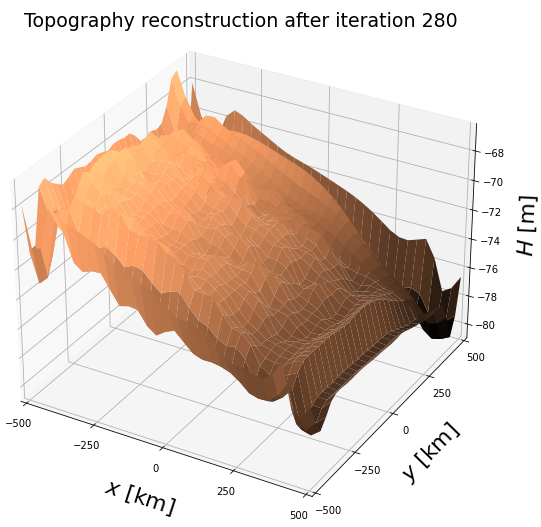}}
    \subfigure{\includegraphics[width=0.33\textwidth]{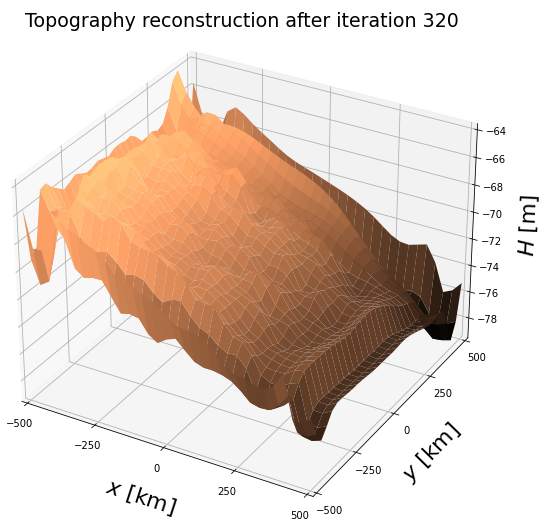}}\\
    \subfigure{\includegraphics[width=0.33\textwidth]{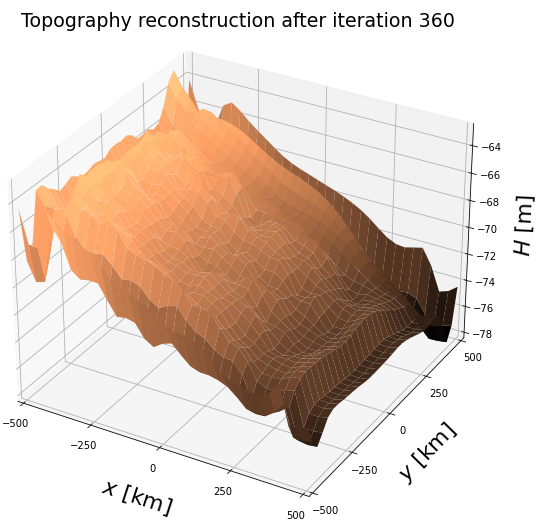}}
    \subfigure{\includegraphics[width=0.33\textwidth]{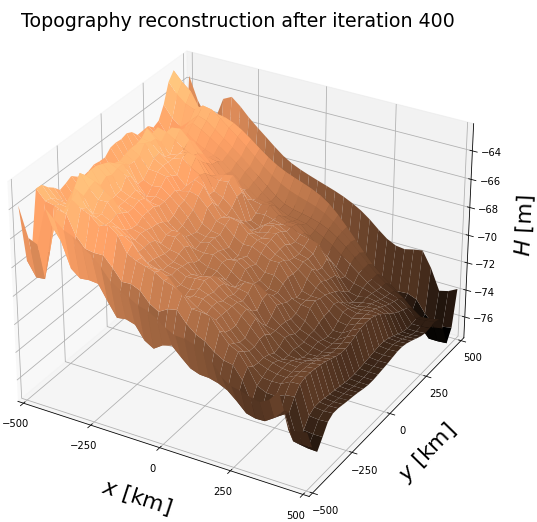}}
    \subfigure{\includegraphics[width=0.33\textwidth]{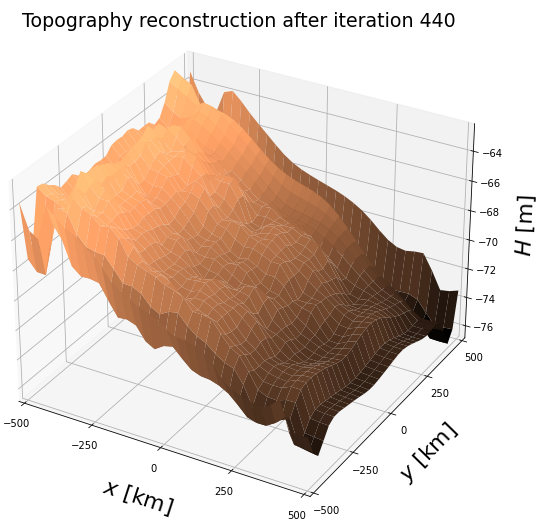}}
\end{figure}
\begin{figure}[ht]
    \subfigure{\includegraphics[width=0.33\textwidth]{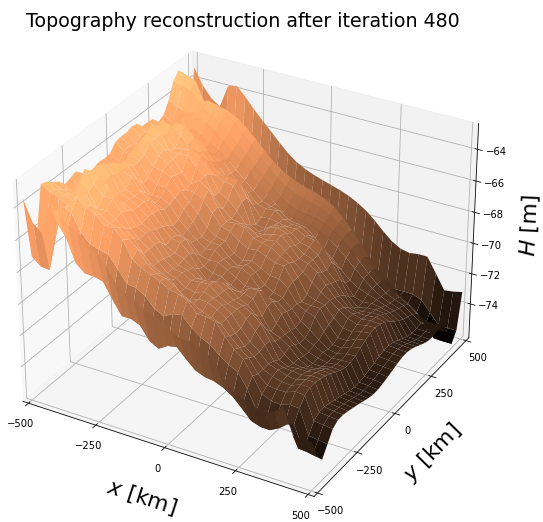}}
    \subfigure{\includegraphics[width=0.33\textwidth]{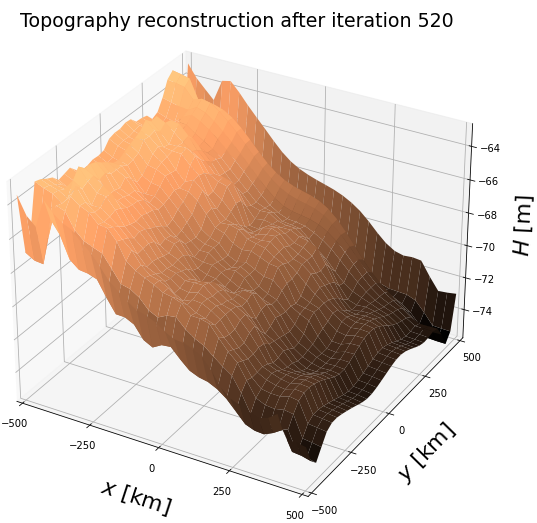}}
    \subfigure{\includegraphics[width=0.33\textwidth]{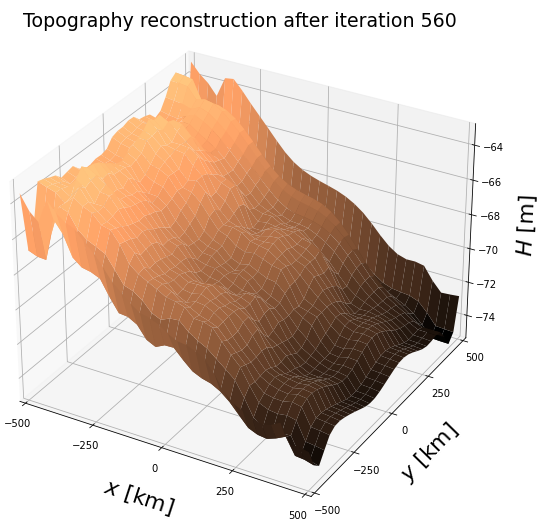}}\\
    \subfigure{\includegraphics[width=0.33\textwidth]{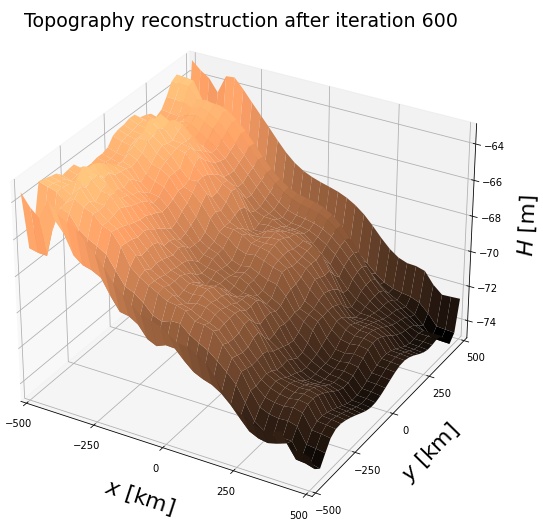}}
    \subfigure{\includegraphics[width=0.33\textwidth]{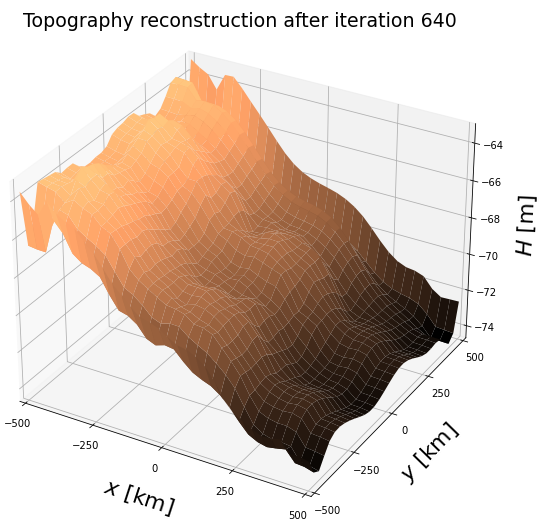}}
    \subfigure{\includegraphics[width=0.33\textwidth]{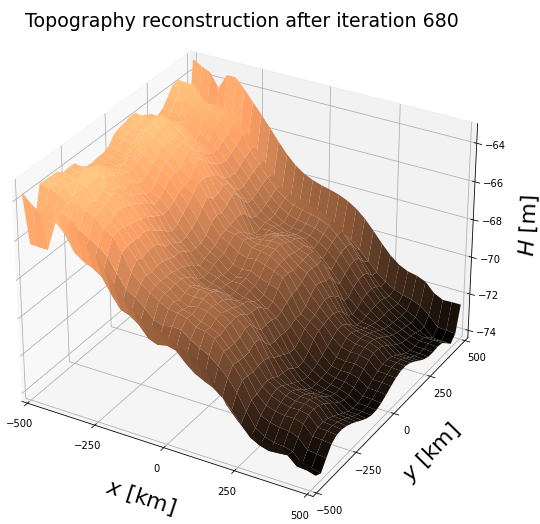}}\\
    \subfigure{\includegraphics[width=0.33\textwidth]{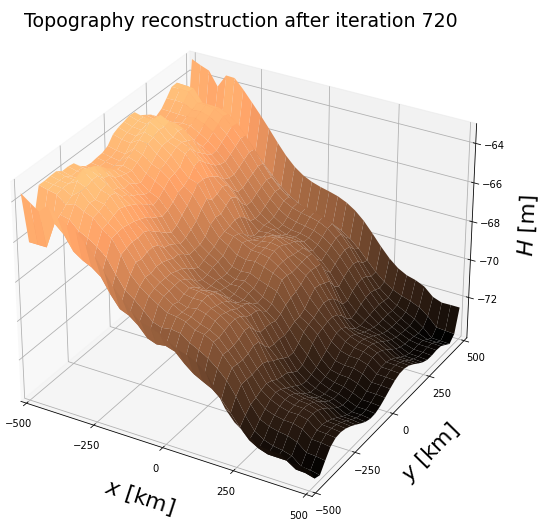}}
    \subfigure{\includegraphics[width=0.33\textwidth]{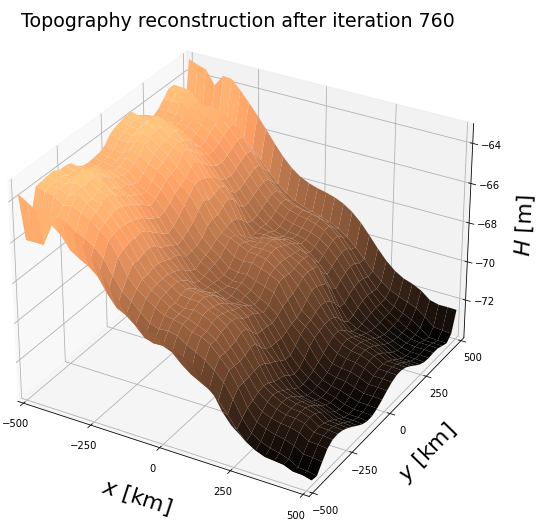}}
    \subfigure{\includegraphics[width=0.33\textwidth]{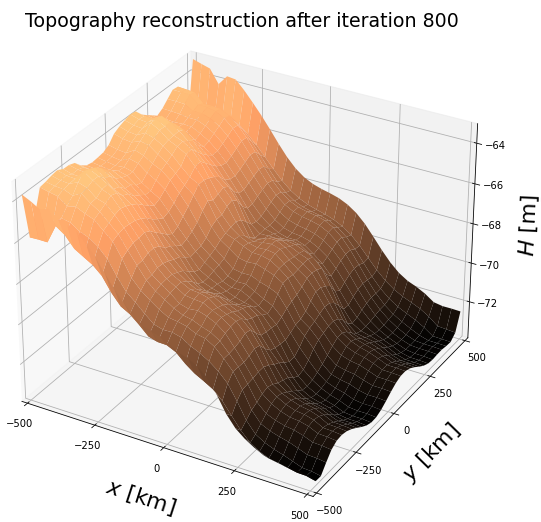}}\\
    \subfigure{\includegraphics[width=0.33\textwidth]{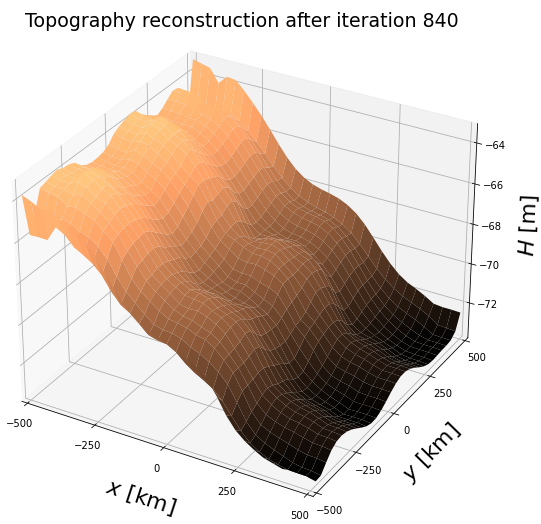}}
    \subfigure{\includegraphics[width=0.33\textwidth]{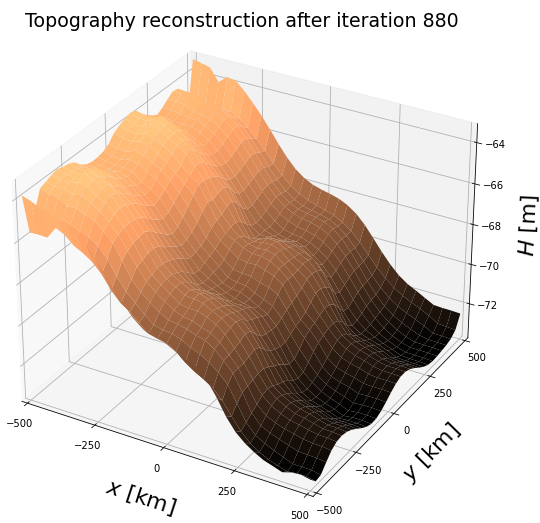}}
    \subfigure{\includegraphics[width=0.33\textwidth]{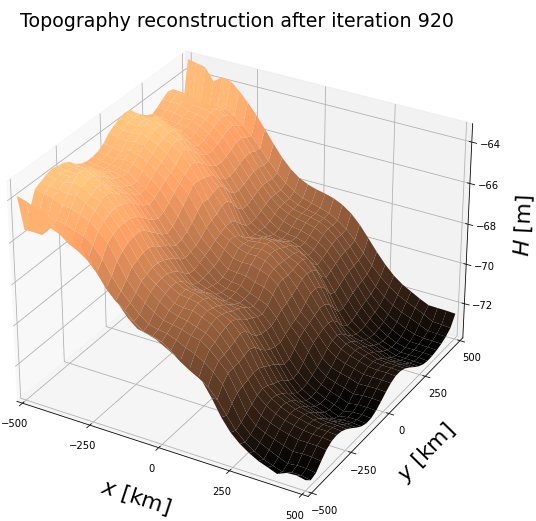}}
\end{figure}
\begin{figure}[ht]
    \subfigure{\includegraphics[width=0.33\textwidth]{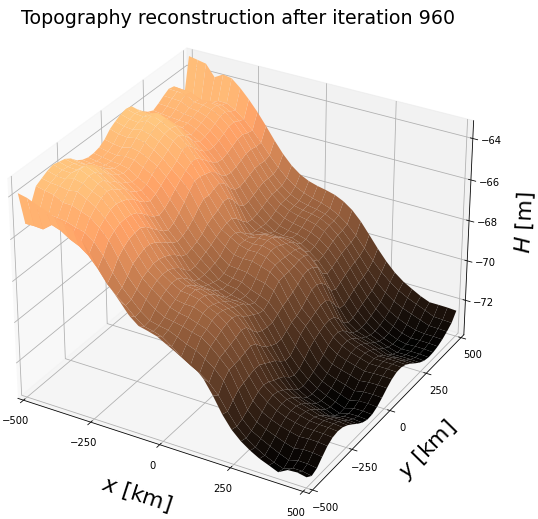}}
    \subfigure{\includegraphics[width=0.33\textwidth]{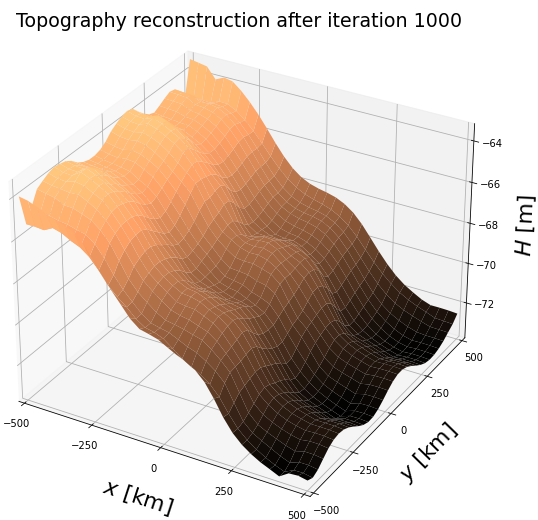}}
    \subfigure{\includegraphics[width=0.33\textwidth]{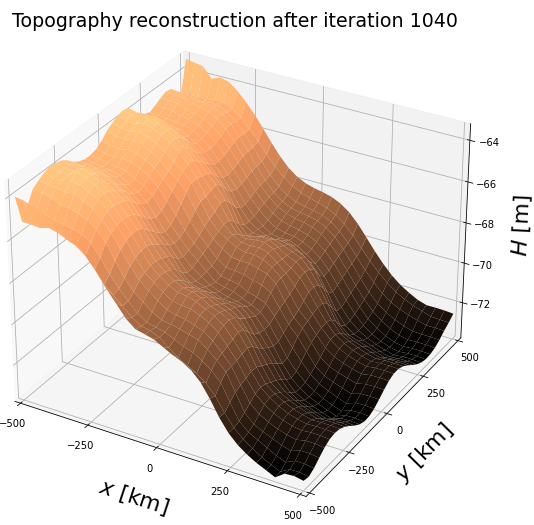}}\\
    \subfigure{\includegraphics[width=0.33\textwidth]{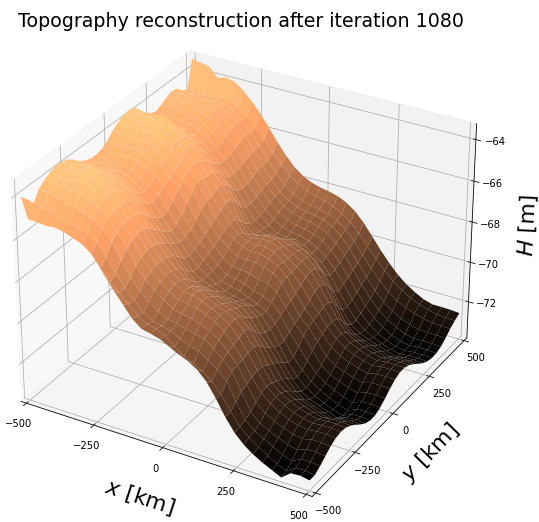}}
    \subfigure{\includegraphics[width=0.33\textwidth]{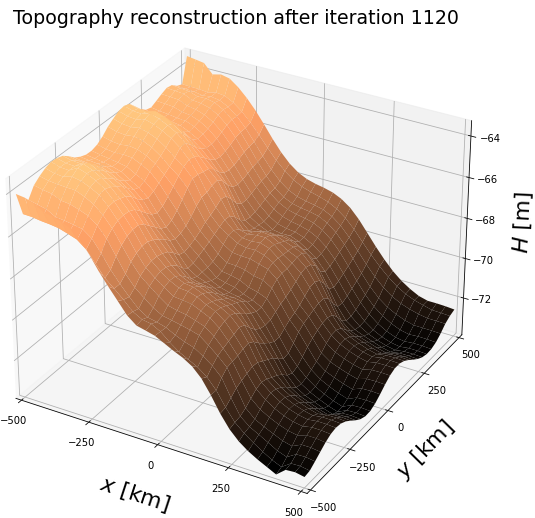}}
    \subfigure{\includegraphics[width=0.33\textwidth]{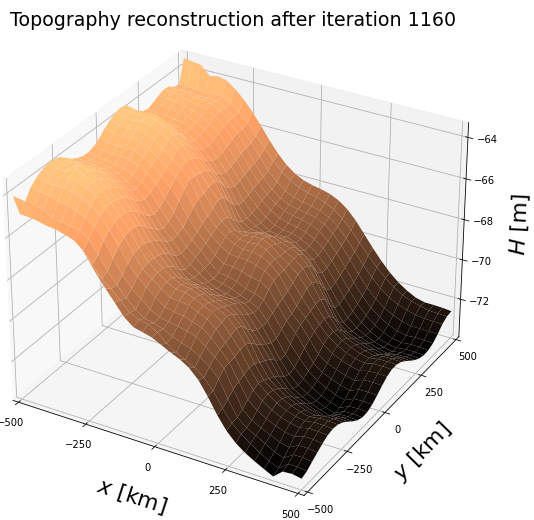}}\\
    \subfigure{\includegraphics[width=0.33\textwidth]{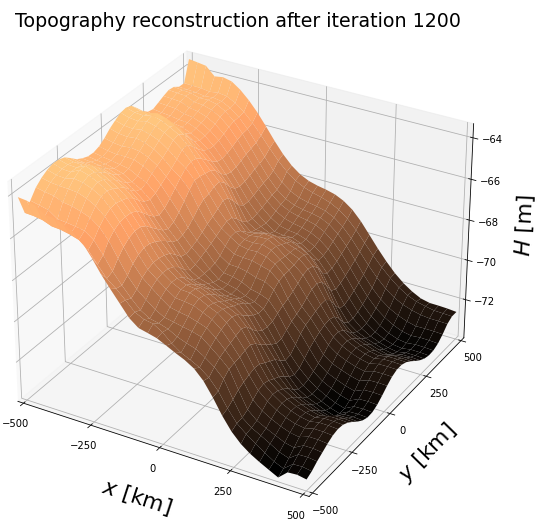}}
    \subfigure{\includegraphics[width=0.33\textwidth]{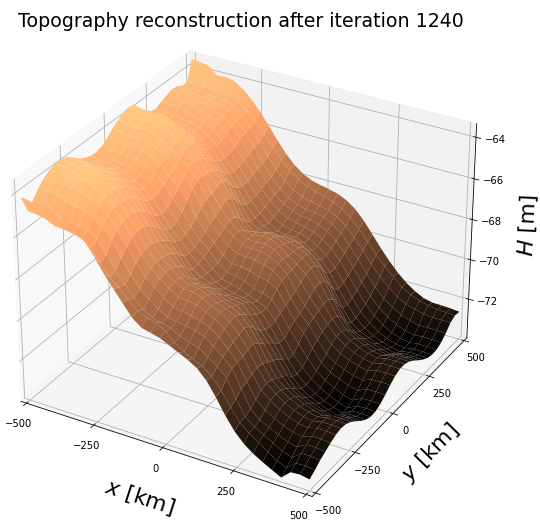}}
    \subfigure{\includegraphics[width=0.33\textwidth]{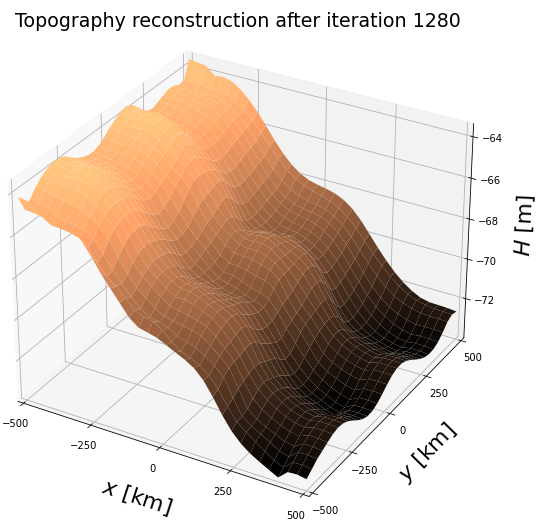}}\\
    \subfigure{\includegraphics[width=0.33\textwidth]{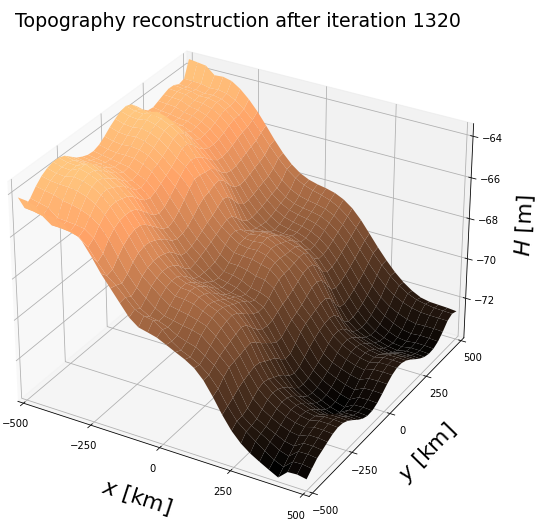}}
    \subfigure{\includegraphics[width=0.33\textwidth]{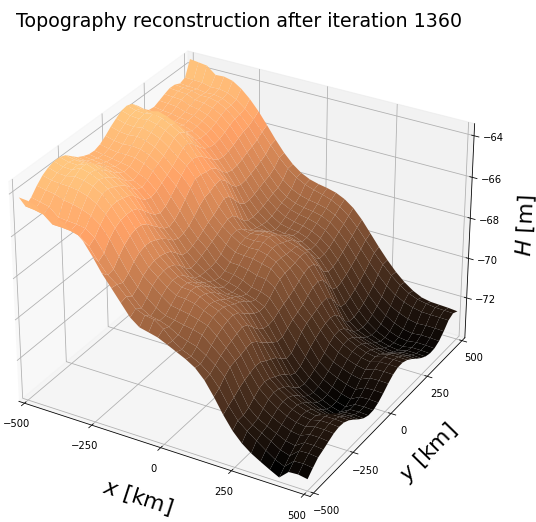}}
    \subfigure{\includegraphics[width=0.33\textwidth]{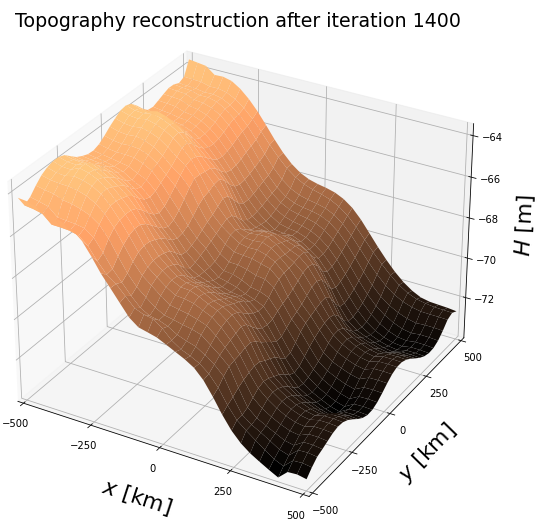}}\\
\end{figure}

\end{document}